\pdfoutput=1
\documentclass[10pt,twocolumn,letterpaper]{article}

\usepackage{caption}
\usepackage{cvpr}
\usepackage{times}
\usepackage{epsfig}
\usepackage{graphicx}
\usepackage{amsmath}
\usepackage{amssymb}
\usepackage{subcaption}
\usepackage[percent]{overpic}
\usepackage{placeins}
\usepackage[table]{xcolor}
\usepackage{enumitem}

\usepackage{colortbl}
\definecolor{gray}{rgb}{0.8,0.8,0.8}
\definecolor{lightblue}{RGB}{176, 224, 230}

\newcommand\crule[3][black]{\textcolor{#1}{\rule{#2}{#3}}}

\cvprfinalcopy

\def\cvprPaperID{2227}

\ifcvprfinal\fi

\begin{document}

%%%%%%%%%%%%%%%%%%%%%%%%%%%%%%%%%%%%%%%%%%%%%%%%%%%%%%%%%%%%%%%%%%%%%%%%%%%%%%%%%%%%%%%%%%%%%%%%%%%
%%%%%%%%% TITLE %%%%%%%%%%%%%%%%%%%%%%%%%%%%%%%%%%%%%%%%%%%%%%%%%%%%%%%%%%%%%%%%%%%%%%%%%%%%%%%%%%%
%%%%%%%%%%%%%%%%%%%%%%%%%%%%%%%%%%%%%%%%%%%%%%%%%%%%%%%%%%%%%%%%%%%%%%%%%%%%%%%%%%%%%%%%%%%%%%%%%%%
\title{YouTube-BoundingBoxes: A Large High-Precision\\
Human-Annotated Data Set for Object Detection in Video}

\author{Esteban Real\\
Google Brain\\
{\tt\small ereal@google.com}
\and
Jonathon Shlens\\
Google Brain\\
{\tt\small shlens@google.com}
\and
Stefano Mazzocchi\\
Google Research\\
{\tt\small stefanom@google.com}
\and
Xin Pan\\
Google Brain\\
{\tt\small xpan@google.com}
\and
Vincent Vanhoucke\\
Google Brain\\
{\tt\small vanhoucke@google.com}
}

\maketitle

%%%%%%%%%%%%%%%%%%%%%%%%%%%%%%%%%%%%%%%%%%%%%%%%%%%%%%%%%%%%%%%%%%%%%%%%%%%%%%%%%%%%%%%%%%%%%%%%%%%
%%%%%%%%%%%%%%%%%%%%%%%%%%%%%%%%%%%%%%%%%%%%%%%%%%%%%%%%%%%%%%%%%%%%%%%%%%%%%%%%%%%%%%%%%%%%%%%%%%%
%%%%%%%%% ABSTRACT %%%%%%%%%%%%%%%%%%%%%%%%%%%%%%%%%%%%%%%%%%%%%%%%%%%%%%%%%%%%%%%%%%%%%%%%%%%%%%%%
%%%%%%%%%%%%%%%%%%%%%%%%%%%%%%%%%%%%%%%%%%%%%%%%%%%%%%%%%%%%%%%%%%%%%%%%%%%%%%%%%%%%%%%%%%%%%%%%%%%
%%%%%%%%%%%%%%%%%%%%%%%%%%%%%%%%%%%%%%%%%%%%%%%%%%%%%%%%%%%%%%%%%%%%%%%%%%%%%%%%%%%%%%%%%%%%%%%%%%%
\begin{abstract}
We introduce a new large-scale data set of video URLs with densely-sampled object bounding box annotations called YouTube-BoundingBoxes (YT-BB). The data set consists of approximately 380,000 video segments about 19s long, automatically selected to feature objects in natural settings without editing or post-processing, with a recording quality often akin to that of a hand-held cell phone camera. The objects represent a subset of the COCO \cite{lin2014microsoft} label set. All video segments were human-annotated with high-precision classification labels and bounding boxes at 1 frame per second. The use of a cascade of increasingly precise human annotations ensures a label accuracy above 95\% for every class and tight bounding boxes. Finally, we train and evaluate well-known deep network architectures and report baseline figures for per-frame classification and localization to provide a point of comparison for future work. We also demonstrate how the temporal contiguity of video can potentially be used to improve such inferences. The data set can be found at https://research.google.com/youtube-bb. We hope the availability of such large curated corpus will spur new advances in video object detection and tracking. 
\end{abstract}

\begin{figure*}
    \captionsetup[subfigure]{skip=0pt}
    \centering
    
    \begin{subfigure}[b]{\textwidth}
        \begin{overpic}[scale=0.76]{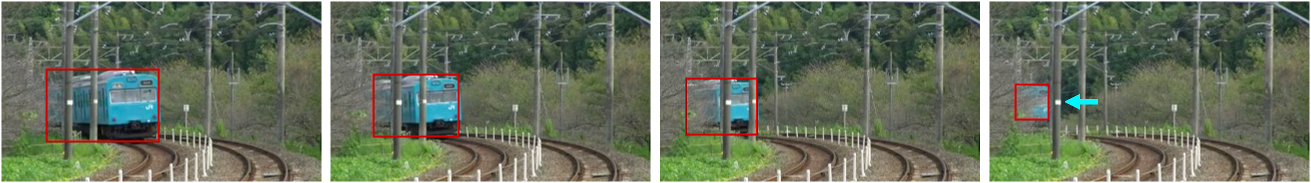}
            \put(18, 11) {\Large \color{red} train}
        \end{overpic}
        \label{fig:example_train}
    \end{subfigure}

    \vspace{-0.5\baselineskip}

    \begin{subfigure}[b]{\textwidth}
        \begin{overpic}[scale=0.76]{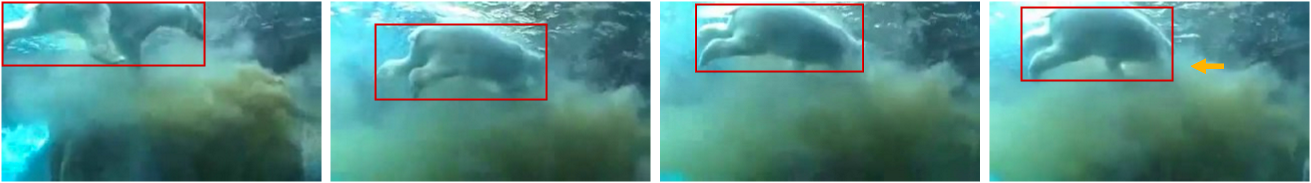}
            \put(18, 11) {\Large \color{red} bear}
        \end{overpic}
        \label{fig:example_bear}
    \end{subfigure}
    
    \vspace{-0.5\baselineskip}

    \begin{subfigure}[b]{\textwidth}
        \begin{overpic}[scale=0.76]{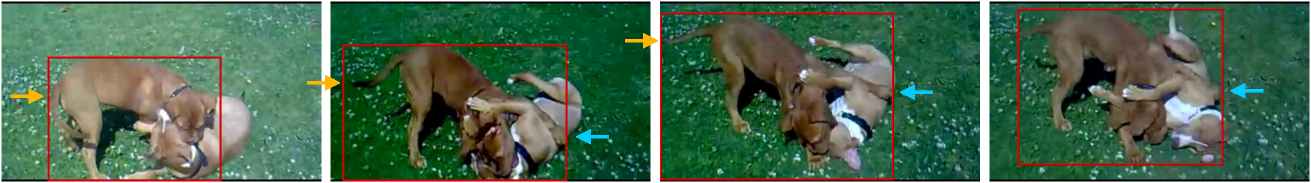}
            \put(19, 11) {\Large \color{red} dog}
        \end{overpic}
        \label{fig:example_dog}
    \end{subfigure}
    
    \vspace{-0.5\baselineskip}

    \begin{subfigure}[b]{\textwidth}
        \begin{overpic}[scale=0.76]{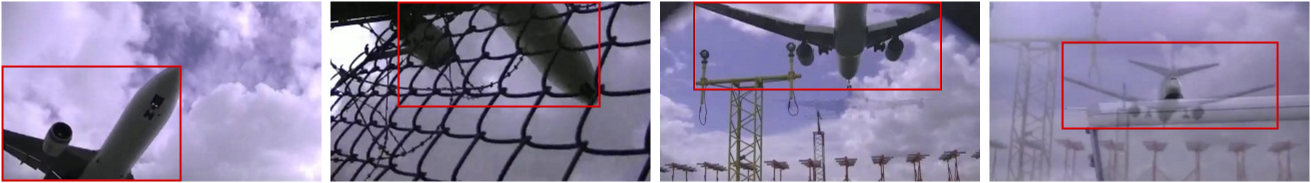}
            \put(14, 11) {\Large \color{red} airplane}
        \end{overpic}
        \label{fig:example_airplane}
    \end{subfigure}
    
    \vspace{-0.5\baselineskip}

    \begin{subfigure}[b]{\textwidth}
        \begin{overpic}[scale=0.76]{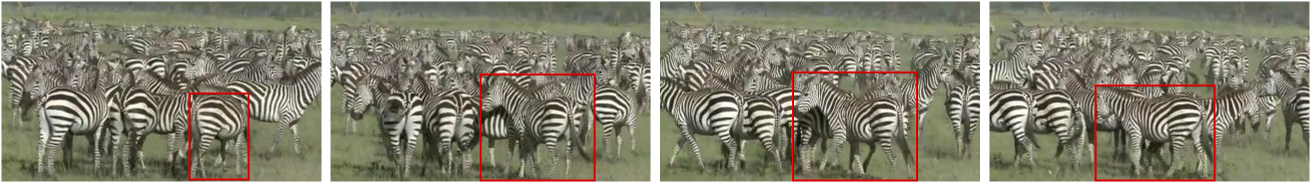}
            \put(17, 11) {\Large \color{red} zebra}
        \end{overpic}
        \label{fig:example_zebra}
    \end{subfigure}
    
    \vspace{-1.0\baselineskip}
    
    \begin{overpic}[scale=0.44]{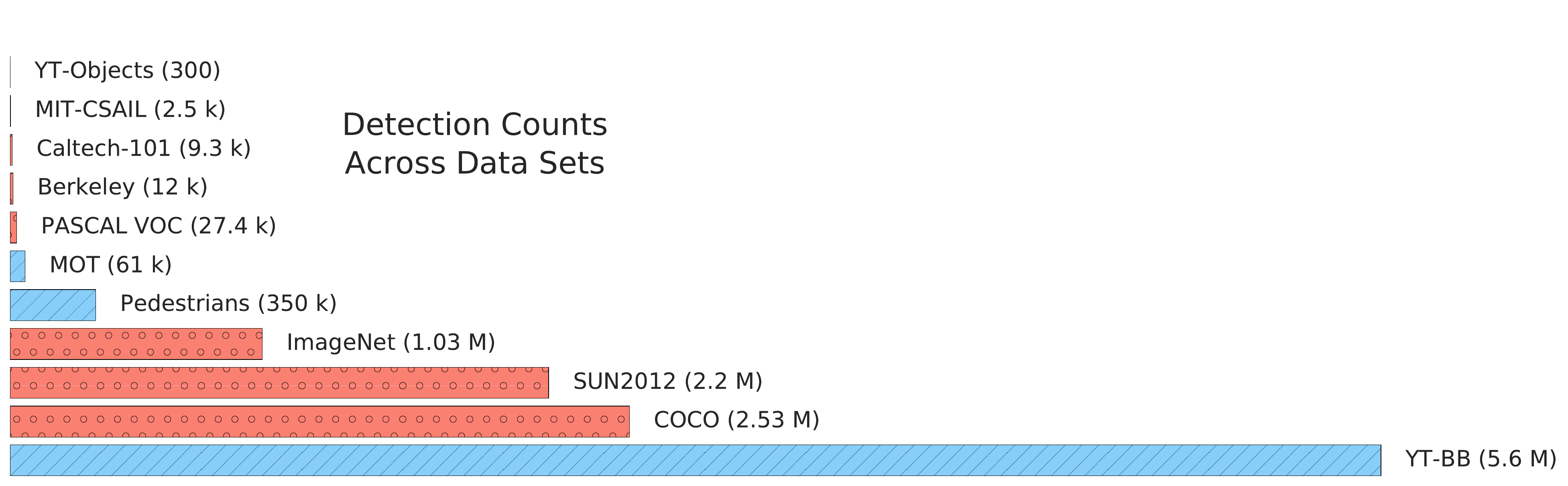}
        \put(53,17){
            \begin{tabular}{l|r|r|r|r}
            \multicolumn{5}{c}{YT-BB Counts and Statistics}  \\[0.2cm]
            & Class. & \cellcolor{lightblue} Boxes & Videos & Motion \\
            \hline
            person & 1.8 M & \cellcolor{lightblue} 1.3 M & 68 k & 0.122 \\
            dog & 560 k & \cellcolor{lightblue} 240 k & 10 k & 0.165 \\
            train & 340 k & \cellcolor{lightblue} 240 k & 8.9 k & 0.072 \\
            \multicolumn{5}{c}{...}  \\
            NONE & 26 k & \multicolumn{1}{|c|}{--} & \multicolumn{1}{|c|}{--} & \multicolumn{1}{|c}{--} \\
            \hline
            TOTAL & 9.5 M & \cellcolor{lightblue} 5.6 M & 240 k & \multicolumn{1}{|c}{--} \\
            \end{tabular}
        }
    \end{overpic}

    \vspace{\baselineskip}

    \caption{IMAGES: Detection examples. Each row shows frames from one video segment. A frame containing an object whose identity can be deduced from other frames is boxed too, as in the last frame of the {\it train example} (blue arrow). Note how only visible parts are included in the box: the orange arrow in the {\it bear example} points to the hidden head. The {\it dog example} illustrates tight bounding boxes tracking the tail (orange arrows) and foot (blue arrows). The {\it airplane example} shows how partial objects are annotated (first frame) and how objects are tracked across changes in perspective, occlusions and camera cuts. Note in the {\it zebra example} how the same object is tracked across multiple frames and other objects of the same class are ignored. BAR CHART: Number of detections/segmentations in various image (red; dots) and video (blue; lines) data sets. The present data set, YT-BB, is at the bottom. TABLE: Human annotation statistics for some classes in YT-BB. The first three columns are counts for: classification annotations, bounding boxes, and unique videos with bounding boxes. Negative classifications are not counted here. The ``Motion'' column shows the RMS of the distance the box travels from one frame to the next, in frame-relative coordinates, to show that the objects exhibit significant motion (see Section \ref{sec:quality_assessment} for details and Supplementary Table \ref{tab:motion} for other measures of movement). The ``NONE'' tag annotates frames that did not have any of the 23 classes.}
    \label{fig:main}
\end{figure*}

%%%%%%%%%%%%%%%%%%%%%%%%%%%%%%%%%%%%%%%%%%%%%%%%%%%%%%%%%%%%%%%%%%%%%%%%%%%%%%%%%%%%%%%%%%%%%%%%%%%
%%%%%%%%%%%%%%%%%%%%%%%%%%%%%%%%%%%%%%%%%%%%%%%%%%%%%%%%%%%%%%%%%%%%%%%%%%%%%%%%%%%%%%%%%%%%%%%%%%%
%%%%%%%%% INTRODUCTION %%%%%%%%%%%%%%%%%%%%%%%%%%%%%%%%%%%%%%%%%%%%%%%%%%%%%%%%%%%%%%%%%%%%%%%%%%%%
%%%%%%%%%%%%%%%%%%%%%%%%%%%%%%%%%%%%%%%%%%%%%%%%%%%%%%%%%%%%%%%%%%%%%%%%%%%%%%%%%%%%%%%%%%%%%%%%%%%
%%%%%%%%%%%%%%%%%%%%%%%%%%%%%%%%%%%%%%%%%%%%%%%%%%%%%%%%%%%%%%%%%%%%%%%%%%%%%%%%%%%%%%%%%%%%%%%%%%%
\section{Introduction}

The exceptional pace of progress in recent years on the tasks of object recognition and detection in still images was enabled by the creation of large-scale, publicly available data sets~\cite{deng2009imagenet, everingham2010pascal, fei2007learning, griffin2007caltech, lin2014microsoft, martin2001database, nene1996columbia, russakovsky2013detecting, torralba2004sharing, xiao2010sun}. These data sets established challenging benchmarks to evaluate new methods for visual object recognition that have substantially improved the state-of-the-art across a broad range of computer vision tasks \cite{he2015deep, krizhevsky2012imagenet, simonyan_vgg, szegedy2016, szegedy2015}.

Open academic challenges paired with open-source recipes have further accelerated the development of the field \cite{caffe_model_zoo, ILSVRC15, tfslim}.
Most notably, systems that perform well on image recognition and object detection may be applied to other computer vision problems in which minimal training data is available \cite{caffe}. Such systems have also become part of larger machine learning pipelines that stretch beyond visual recognition (\eg multi-modal learning \cite{conse,caption}).

The increased speed and memory of modern computing architectures places the research community in a position to aim for comparable results in video, a natural goal for machine perception. The quest for large video data sets, however, has been more elusive. One challenge is that the online corpus of videos is \textit{weakly labeled}, \ie the label information is very noisy \cite{couchpotato}. Sifting through a large sample may therefore require considerable human involvement.

Exacerbating this problem is the recognition that large data sets are necessary to prevent over-fitting of cutting-edge models (\eg \cite{shortsnippets, simonyan}). Although the temporal dimension provides vastly more data, much of the information is redundant due to correlations of pixels across frames. Thus, increasing the data set size is not merely about gathering more sequential frames from a small number of videos. Instead, we need a large, diverse sample of videos. Attaining it requires paying special attention to how the videos are mined. Some of the larger existing vision data sets rely indirectly on aggregate measurements of human preference \cite{pamir}. Consequently, those data sets favor aesthetically pleasing viewpoints of labeled objects. This leads to object recognition systems that are precise but may lack variety in terms of realistic lighting conditions, occlusions or the non-canonical viewpoints often observed in real life. Video may be less prone to some of these biases (especially the viewpoint bias), but a random YouTube sample would still suffer drastically from them. Mining videos with diversity in mind, on the other hand, can address this problem explicitly.

The persistence and temporal consistency of objects present in natural video scenes call for a different kind of labeling, whereby objects of interest are tracked across frames and precisely localized. To this day, there is no human-curated large-scale data set that provides classification and detection annotations for objects of several classes in a wide variety of videos.

This work attempts to address this issue by providing a large body of video annotations with manually curated bounding boxes of objects tracked for relatively long durations on the order of $100$ frames. The size of the data set makes it suitable for training large deep neural networks and explore visio-temporal modeling approaches in a realistic setting.

%%%%%%%%%%%%%%%%%%%%%%%%%%%%%%%%%%%%%%%%%%%%%%%%%%%%%%%%%%%%%%%%%%%%%%%%%%%%%%%%%%%%%%%%%%%%%%%%%%%
%%%%%%%%%%%%%%%%%%%%%%%%%%%%%%%%%%%%%%%%%%%%%%%%%%%%%%%%%%%%%%%%%%%%%%%%%%%%%%%%%%%%%%%%%%%%%%%%%%%
%%%%%%%%% RELATED WORK %%%%%%%%%%%%%%%%%%%%%%%%%%%%%%%%%%%%%%%%%%%%%%%%%%%%%%%%%%%%%%%%%%%%%%%%%%%%
%%%%%%%%%%%%%%%%%%%%%%%%%%%%%%%%%%%%%%%%%%%%%%%%%%%%%%%%%%%%%%%%%%%%%%%%%%%%%%%%%%%%%%%%%%%%%%%%%%%
%%%%%%%%%%%%%%%%%%%%%%%%%%%%%%%%%%%%%%%%%%%%%%%%%%%%%%%%%%%%%%%%%%%%%%%%%%%%%%%%%%%%%%%%%%%%%%%%%%%
\section{Related work}
\label{sec:related_work}

Several video data sets are already available to the community. Below are some of the most relevant, highlighting how they differ from YouTube-BoundingBoxes:
\begin{itemize}[leftmargin=0cm,itemindent=.3cm,labelwidth=\itemindent,labelsep=0cm,align=left,noitemsep,topsep=0pt]
    \item Many data sets such as the HMDB-51 data set \cite{Kuehne11} and the UCF-101 data set \cite{soomro2012ucf101} provide segment-level annotations for a variety of human action categories; richer annotations including fine-grained temporal and localization information were provided as part of the THUMOS \cite{THUMOS15} challenge.
    \item TRECVID \cite{2016trecvidawad} is a yearly set of competitions centered on video retrieval and indexing, hosting a variety of video data sets. For 2016, they provide a localization test set with 1000 videos annotated with bounding boxes for 10 classes; each video may or may not contain a box.
    \item VOT \cite{kristan2015visual} and MOT \cite{MOTChallenge2015} are yearly visual object tracking challenges with associated data sets. These are small and extensively curated in order to provide controlled frequencies of various common difficulties for object tracking (such as a occlusions, illumination changes or size changes).
    \item The Sports-1M data set \cite{KarpathyCVPR14,youtube_sports} consists of segment-level annotations for a variety of sports, with temporal localization. % VV: I removed 'manually annotated'
    \item The YouTube-Objects data set \cite{prest2012learning,youtube_objects} consists of a number of frames queried from YouTube with a few hundred curated bounding box annotations.
    \item The Caltech Pedestrian Detection data set \cite{dollar2009pedestrian} consists of 350,000 bounding boxes of pedestrians annotated from a vehicle driving through an urban environment.
    \item The YouTube-8M data set \cite{1609.08675} consists of a very large set of frame-level, automatically generated annotations of YouTube videos. The labels were generated using state-of-the-art deep networks to classify thousands of possible entities.
    \item ImageNet 2015 \cite{ILSVRC15} has a video object detection data set with 5,400 videos.
\end{itemize}

Still-image detection data sets are larger and more abundant. They vary in detail from bounding boxes (Caltech-101 \cite{fei2007learning}, MIT-CSAIL \cite{torralba2004sharing}, ImageNet \cite{deng2009imagenet, russakovsky2013detecting}, PASCAL VOC \cite{everingham2015pascal, everingham2010pascal}, SUN2012 \cite{xiao2014sun, xiao2010sun}) to pixel-level segmentations (Berkeley Segmentation Data Set \cite{martin2001database}, Caltech-101 \cite{fei2007learning}, PASCAL VOC \cite{everingham2015pascal, everingham2010pascal}, Microsoft COCO \cite{lin2014microsoft}).

The bar chart in Figure \ref{fig:main} puts our data set in context: YT-BB is the largest human-annotated detection data set in existence so far. Specifically for the case of video, it exceeds other data sets in size by more than an order of magnitude.

%%%%%%%%%%%%%%%%%%%%%%%%%%%%%%%%%%%%%%%%%%%%%%%%%%%%%%%%%%%%%%%%%%%%%%%%%%%%%%%%%%%%%%%%%%%%%%%%%%%
%%%%%%%%%%%%%%%%%%%%%%%%%%%%%%%%%%%%%%%%%%%%%%%%%%%%%%%%%%%%%%%%%%%%%%%%%%%%%%%%%%%%%%%%%%%%%%%%%%%
%%%%%%%%% METHODS %%%%%%%%%%%%%%%%%%%%%%%%%%%%%%%%%%%%%%%%%%%%%%%%%%%%%%%%%%%%%%%%%%%%%%%%%%%%%%%%%
%%%%%%%%%%%%%%%%%%%%%%%%%%%%%%%%%%%%%%%%%%%%%%%%%%%%%%%%%%%%%%%%%%%%%%%%%%%%%%%%%%%%%%%%%%%%%%%%%%%
%%%%%%%%%%%%%%%%%%%%%%%%%%%%%%%%%%%%%%%%%%%%%%%%%%%%%%%%%%%%%%%%%%%%%%%%%%%%%%%%%%%%%%%%%%%%%%%%%%%
\section{Methods}

%%%%%%%%%%%%%%%%%%%%%%%%%%%%%%%%%%%%%%%%%%%%%%%%%%%%%%%%%%%%%%%%%%%%%%%%%%%%%%%%%%%%%%%%%%%%%%%%%%%
%%%%%%%%% DATA MINING %%%%%%%%%%%%%%%%%%%%%%%%%%%%%%%%%%%%%%%%%%%%%%%%%%%%%%%%%%%%%%%%%%%%%%%%%%%%%
%%%%%%%%%%%%%%%%%%%%%%%%%%%%%%%%%%%%%%%%%%%%%%%%%%%%%%%%%%%%%%%%%%%%%%%%%%%%%%%%%%%%%%%%%%%%%%%%%%%
\subsection{Data mining}
\label{sec:data_mining}

In order to provide a low entry-bar to video for researchers that have models pre-trained on static images, we chose as our labels 23 classes that form a subset of the detection classes in the COCO data set \cite{lin2014microsoft}. Due to its particular importance, we included the ``person'' class and gave it preferential treatment in terms of total volume and in terms of how the videos were mined (details below). The other classes are all common objects or animals (first column in Supplementary Table \ref{tab:classification_counts}).

Many academic data sets are made artificially easy compared to real-world problem settings because they have a closed set of labels to chose from, whereas most data collected ``in the wild'' can't be expected to correspond to a well-defined category. This is particularly important for detection and localization tasks. To directly address this problem, we added a ``NONE'' class that marks frames that do not have any of the 23 object classes.

We sampled public YouTube videos and used object-agnostic signals to reduce the set obtained to a size suitable for human annotation. We calculated an estimate of the entropy across frames and removed those below a particular threshold, reducing the frequency of slide shows and other videos with minimal motion. Requiring that videos have fewer than $100$ views notably reduced the number of professionally edited clips. More generally, this limit on the view count helps protect against the bias of a plain internet search result, which would yield preferentially videos that are likable (good lighting, centered characters, stable cameras, \etc) Finally, a camera-cut detector helped remove videos that had unusually short scenes, which are indicative of a high degree of post-processing. We then split the remaining videos into short non-overlapping clips (mean length = 18.7s, sd = 1.00s). All these restrictions together resulted in a collection of video segments typical of what a hand-held camera would record in a natural setting.

This data mining procedure proved satisfactory for the ``person'' class, but was too inefficient for classes that occur infrequently in the YouTube corpus. To compensate, we ran image classifiers at 1 frame per second across our video sample. We retained the top 1 million videos, discarding those deemed by the classifiers as too unlikely to contain any of our 23 classes\footnote{We intentionally ran the image classifiers at low thresholds for each class in order to avoid the pitfall of selecting easy-to-classify examples. Specifically, we ranked candidate segments according to the confidence of the classifiers and set the threshold for a given image classifier to operate at low rates of precision, as judged by one-off experiments. Our selection of threshold had the goal of leaving plenty of work for human annotators to do as far as discriminating the presence or absence of each class.}. When possible, we exploited the WordNet hierarchy \cite{wordnet} to associate multiple fine-grain image labels with a given class label in YT-BB.

The ``person'' class is especially important. In particular, the research community has a vested interest in detecting people in videos. While our initial approach of sampling random YouTube videos for this class may provide a fairly unbiased data set, it may also produce one that is too homogeneous. The most popular videos (like a recent music album) are not a problem because they were removed by the initial object-agnostic filters. However, there is still a class of videos that may be {\it filmed} very frequently even if they are not viewed many times by other users. This would include the sort of things most of us care about, like birthday parties, graduation ceremonies, and the like. As an attempt to compensate for that, we enriched our random sample of ``person'' videos with a comparable yet smaller number of videos mined from entities that correlated with ``person'' well (``person'', ``bicycle'', ``crowd'' and--surprisingly--``elephant'' are examples). Finally, we intentionally mined a disproportionately large number of videos for this class with the goal that the ``person'' subset of our data set may stand on its own.

Balancing the time spent on human annotation and the yield required focusing on segments that usually contain only one class. We felt this was preferable to the huge sacrifice in volume that would have been necessary to label segments containing multiple classes. Namely, mining for videos with several classes results in a much lower yield and the alternative of mining them with higher recall produces too many false negatives which in turn increases the human annotation time too much.

%%%%%%%%%%%%%%%%%%%%%%%%%%%%%%%%%%%%%%%%%%%%%%%%%%%%%%%%%%%%%%%%%%%%%%%%%%%%%%%%%%%%%%%%%%%%%%%%%%%
%%%%%%%%% HUMAN ANNOTATIONS %%%%%%%%%%%%%%%%%%%%%%%%%%%%%%%%%%%%%%%%%%%%%%%%%%%%%%%%%%%%%%%%%%%%%%%
%%%%%%%%%%%%%%%%%%%%%%%%%%%%%%%%%%%%%%%%%%%%%%%%%%%%%%%%%%%%%%%%%%%%%%%%%%%%%%%%%%%%%%%%%%%%%%%%%%%
\subsection{Human annotations}
\label{sec:human_annotation}

Like other large data sets before us \cite{deng2009imagenet, lin2014microsoft}, we used human annotation pipelines to label our data. As in \cite{lin2014microsoft}, we set up a cascade of stages that successively refine the quality of the results. This strategy is standard \cite{chen2011opportunities} and has been found to improve results \cite{bernstein2015soylent}. We used four stages:
\begin{enumerate}[leftmargin=0cm,itemindent=.4cm,labelwidth=\itemindent,labelsep=0cm,align=left,noitemsep,topsep=0pt]
  \item Five frames from each ($\sim 19$ s) video segment, evenly sampled in time, were simultaneously presented to one human rater (i.e. annotator), who had to determine whether a specific class was present in any of them. Negative segments were discarded.
  \item Each full segment was presented to three different annotators as a ``movie-roll'', sampled at $1$ frame per second. The annotators had to indicate whether the class was present in each frame. The majority vote produced our (intermediate) classification data set. To find frames for the ``NONE'' class, we asked the raters explicitly about each of the 23 classes to ensure they were absent. Such annotations are precise but very time-consuming, and so the frequency of the ``NONE'' class is limited (see table in Figure \ref{fig:main} or Supplementary Table \ref{tab:classification_counts}). Segments with at least one positive frame for a given class were used in stages 3 and 4.
  \item For each segment, a single human annotator overlaid a bounding box tightly around an object of the given class in each of the frames in the segment, at $1$ frame per second. Every appearance of a single object was annotated throughout the segment. (Other objects of the same class were to be ignored). The annotator also had the option of assigning an \textit{absent tag} to a frame if the object could not be seen there. To resolve corner cases, they followed the guidelines in Supplementary Section \ref{sec:box_guidelines}, which address issues of box tightness, partial objects, occlusion, \etc. Incidentally, these rules may help readers clarify peculiarities of our data set.
  \item Each annotation from stage 3 was verified by one (training and validation data sets) or three (testing data set) different human annotators. Boxes or absent-tags with negative majority votes were discarded.
\end{enumerate}

We employed {\it Amazon Mechanical Turk} for the first two stages of human annotation as in \cite{deng2009imagenet, lin2014microsoft}. This allowed for quick progress \cite{buhrmester2011amazon}, yet suffered from the widely known drawbacks of crowd-sourcing, including difficulty motivating raters and poor quality of individual annotations \cite{ipeirotis2010quality}. This can be partly curbed through replication \cite{paritosh2012human, sheng2008get}, as we did in stage 2. While there exist more sophisticated methods for analyzing replicated labels \cite{paritosh2012human, sheng2008get, waterhouse2013pay}, we opted for the majority vote because it was simple, we only had 3 labels per example, and the data was going to be further refined by stages 3 and 4 anyway. In order to harness the benefits of annotator training \cite{dow2011shepherding}, for stages 3 and 4 we switched to our internal human annotation system. We employed human raters that read a written manual describing the task in detail and went over it during class sessions. During the annotation process, they were also able to escalate questions when they felt unsure about corner cases.

Another important aspect of human computation is the annotator's interaction with the data. We designed user interfaces (Supplementary Section \ref{sec:human_annotation_user_interfaces}) in keeping with the principle that ``the simpler, the better'' \cite{finnerty2013keep}. Especially for {\it Mechanical Turk} tasks, it was important to phrase the questions well, striking a balance between reducing ambiguity and keeping the operator's attention (details in Supplementary Section \ref{sec:attention_span}).

To fine tune the pipeline, we frequently inspected the data by eye. Stages 1 and 2 were finalized only after the resulting classification accuracy was estimated to be above $0.95$ for each class. Stages 3 and 4 were optimized by giving feedback to annotators based on (i) examples where stage 4 showed the most disagreement and (ii) randomly sampled examples. As we increased the size of the annotation batches, rarer examples of type (i) appeared. The quality seemed to improve with annotator experience too, so the validation and testing subsets were done last. The resulting quality after stage 4 is discussed below.

%%%%%%%%%%%%%%%%%%%%%%%%%%%%%%%%%%%%%%%%%%%%%%%%%%%%%%%%%%%%%%%%%%%%%%%%%%%%%%%%%%%%%%%%%%%%%%%%%%%
%%%%%%%%%%%%%%%%%%%%%%%%%%%%%%%%%%%%%%%%%%%%%%%%%%%%%%%%%%%%%%%%%%%%%%%%%%%%%%%%%%%%%%%%%%%%%%%%%%%
%%%%%%%%% RESULTS AND QUALITY ASSESSMENT %%%%%%%%%%%%%%%%%%%%%%%%%%%%%%%%%%%%%%%%%%%%%%%%%%%%%%%%%%
%%%%%%%%%%%%%%%%%%%%%%%%%%%%%%%%%%%%%%%%%%%%%%%%%%%%%%%%%%%%%%%%%%%%%%%%%%%%%%%%%%%%%%%%%%%%%%%%%%%
%%%%%%%%%%%%%%%%%%%%%%%%%%%%%%%%%%%%%%%%%%%%%%%%%%%%%%%%%%%%%%%%%%%%%%%%%%%%%%%%%%%%%%%%%%%%%%%%%%%
\section{Results}
\label{sec:results}

\subsection{Data set size}

\begin{figure}[t]
\begin{center}
   \includegraphics[width=0.95\linewidth]{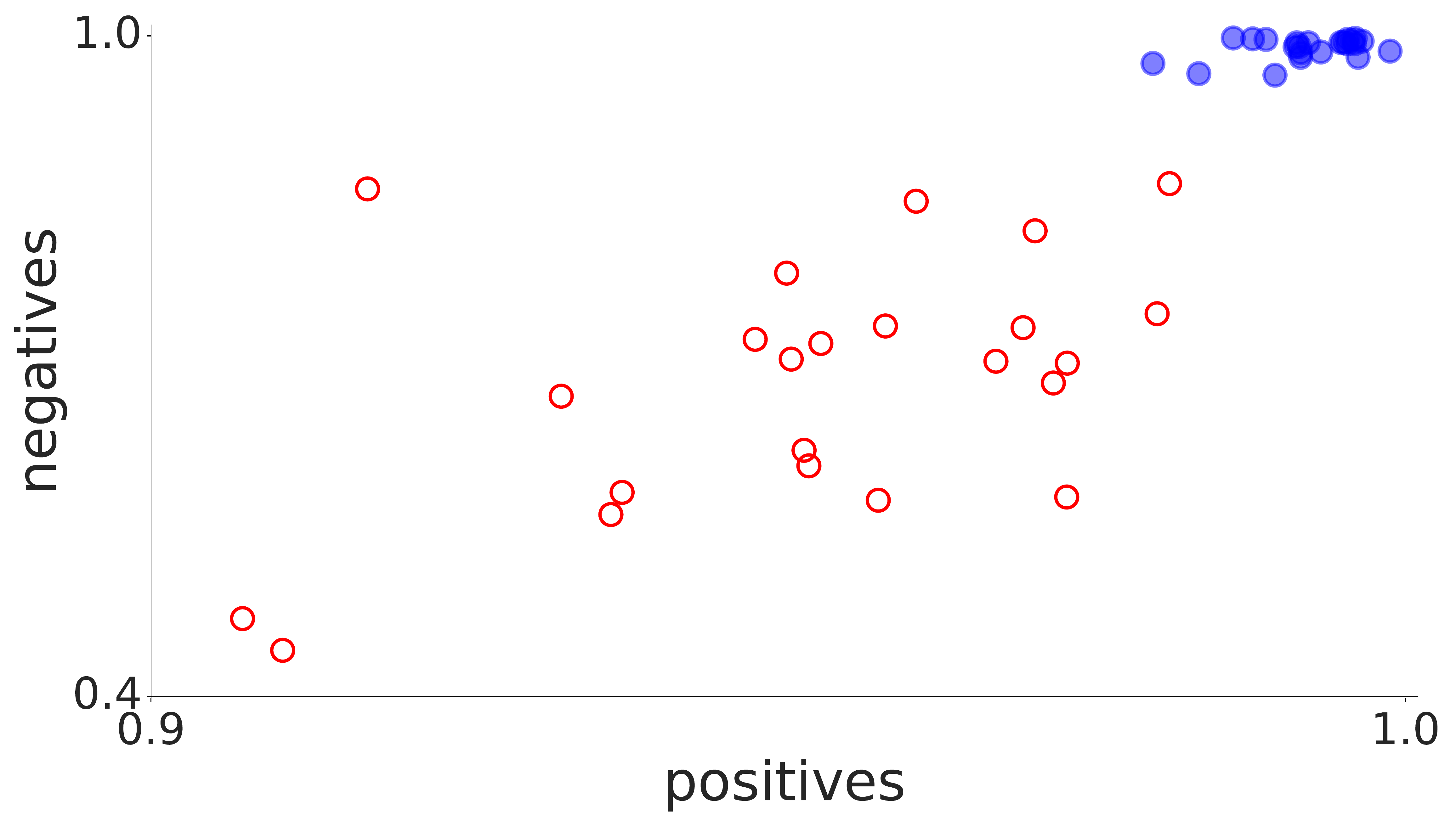}
\end{center}
   \caption{Unanimous fraction per class. The fraction of answers for which the operators voted unanimously, for classification (empty red circles) and detection (filled blue circles). The axes distinguish positives from negatives. For the classifications, this is based on the frame-level annotations, which were done by three human annotators. For the detections, the bounding box drawing stage was done by only one annotator, so this is based on the verification stage of the testing data set, which was done by three annotators too.}
\label{fig:fraction_correct}
\end{figure}

This process yielded a data set of 5.6 million frames annotated with bounding boxes from 240,000 unique YouTube videos. We also provide additional absent detection tags in $1$ million frames from 55,000 unique videos. A superset of those videos contain classification annotations too: $9.6$ million positives and $1$ million negatives, with a similar distribution over unique videos. This is presented in detail in Supplementary Tables \ref{tab:classification_counts} and \ref{tab:detection_counts} and succinctly in the table in Figure \ref{fig:main}, together with some examples.

\subsection{Quality assessment}
\label{sec:quality_assessment}

In order to assess the quality of the classifications, we measured the fraction of answers that were unanimous. Stage 1 strongly biases our sample toward positives (Section \ref{sec:human_annotation}), which results in a higher fraction of false negative classifications. The use of untrained, unvetted raters also seriously reduces the accuracy of the answers (Figure \ref{fig:fraction_correct}). While this could be improved, our main goal of classifying the videos is to filter them in order to draw the bounding boxes, and so we did not optimize stages 1 and 2 further. Nevertheless, we make them available in our data set.

For the detections, we asked raters to verify the bounding boxes (or the absent tag). The frequency of correct verifications is an indication of the quality of the boxes. By this measure, each class had at least $98\%$ correct bounding boxes and at least $98\%$ correct absent-tags. In the case of the testing data set (for which we employed three raters), we can consider the harsher criterion of requiring a unanimously correct verification vote (instead of just a majority-correct verification vote): this gave that both, boxes and absent-tags, are still at least $98\%$ correct for most classes and all classes are above $95\%$ correct (Figure \ref{fig:fraction_correct}).

Annotation quality aside, a concern is that the objects in the videos exhibit movement. Otherwise, the data set would be equivalent to static images. We measured the RMS of the distance the center of the bounding boxes travels from one frame to the next and found that there is indeed significant motion. A few values are quoted in the table in Figure \ref{fig:main}. Results for all classes can be found in Supplementary Table \ref{tab:motion}. Other statistics are also listed there, such as the fractional size change of the box per second (min: 7.2\% for train, max: 19\% for skateboard), how often it enters and exists the field of view, how much area it covers and how frequently it is present.

\subsection{Data set splits}

The final annotations were split into training, validation, and testing subsets, as is standard for machine learning applications. The validation and testing subsets comprise $10\%$ of the total, and this fraction is constant across classes. The splits were done such that no YouTube video can straddle two subsets. Part of the testing subset will be withheld in order to provide a quality measure for future public challenges based on YT-BB.

%%%%%%%%%%%%%%%%%%%%%%%%%%%%%%%%%%%%%%%%%%%%%%%%%%%%%%%%%%%%%%%%%%%%%%%%%%%%%%%%%%%%%%%%%%%%%%%%%%%
%%%%%%%%%%%%%%%%%%%%%%%%%%%%%%%%%%%%%%%%%%%%%%%%%%%%%%%%%%%%%%%%%%%%%%%%%%%%%%%%%%%%%%%%%%%%%%%%%%%
%%%%%%%%% BASELINE MODELS %%%%%%%%%%%%%%%%%%%%%%%%%%%%%%%%%%%%%%%%%%%%%%%%%%%%%%%%%%%%%%%%%%%%%%%%%
%%%%%%%%%%%%%%%%%%%%%%%%%%%%%%%%%%%%%%%%%%%%%%%%%%%%%%%%%%%%%%%%%%%%%%%%%%%%%%%%%%%%%%%%%%%%%%%%%%%
%%%%%%%%%%%%%%%%%%%%%%%%%%%%%%%%%%%%%%%%%%%%%%%%%%%%%%%%%%%%%%%%%%%%%%%%%%%%%%%%%%%%%%%%%%%%%%%%%%%
\section{Baseline models}
\label{sec:baseline_models}

We measured the performance on YT-BB of image classification and object detection models trained on the COCO data set and vice-versa. This is possible because YT-BB's labels are a subset of COCO's, and both data sets classify and localize objects. The goal of this analysis two-fold: (1) to establish the relative difficulty of either task on the two datasets and (2) to provide a point of comparison for future network architectures.

%%%%%%%%%%%%%%%%%%%%%%%%%%%%%%%%%%%%%%%%%%%%%%%%%%%%%%%%%%%%%%%%%%%%%%%%%%%%%%%%%%%%%%%%%%%%%%%%%%%
%%%%%%%%% INCEPTION BASELINE %%%%%%%%%%%%%%%%%%%%%%%%%%%%%%%%%%%%%%%%%%%%%%%%%%%%%%%%%%%%%%%%%%%%%%
%%%%%%%%%%%%%%%%%%%%%%%%%%%%%%%%%%%%%%%%%%%%%%%%%%%%%%%%%%%%%%%%%%%%%%%%%%%%%%%%%%%%%%%%%%%%%%%%%%%
\subsection{Image classification}

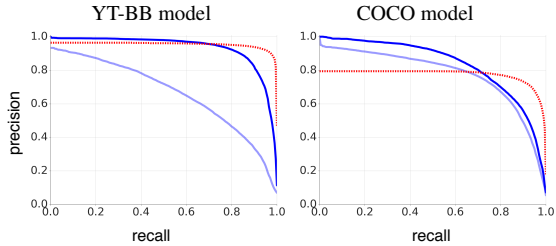
\begin{figure}[t]
\begin{center}
    \begin{overpic}[scale=0.38]{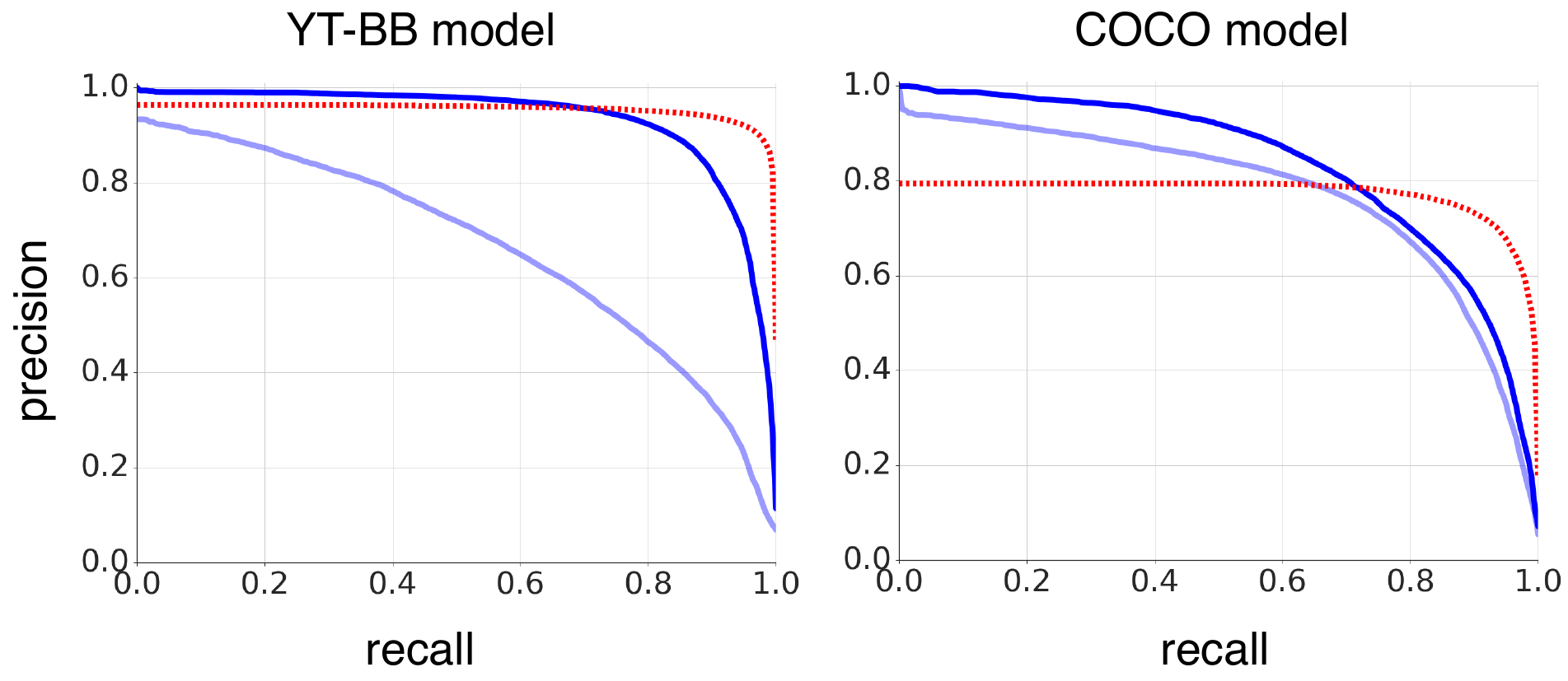}
       \put(17, 39) {\crule[white]{1.7cm}{0.5cm}}
       \put(16, 41) {\footnotesize YT-BB model}
       \put(65, 39) {\crule[white]{1.7cm}{0.5cm}}
       \put(64, 41) {\footnotesize COCO model}
    \end{overpic}
\end{center}
   \caption{Baseline comparison on image classification. Mean precision-recall curve across 23 classes trained on the YT-BB (left) and the COCO~\cite{lin2014microsoft} data sets (right). The dark line is the test set for the same data set; the light line is the test set for the opposite data set. The red dashed line is the evaluation where the classification scores are averaged across a video segment.}
\label{fig:baseline-classification}
\end{figure}

\begin{table}
\begin{center}
\begin{tabular}{ |c|c|c|c|c| } 
 \hline
 Train. data & Eval. data & smooth? & mAP & AUC \\
 \hline
 COCO & COCO & - & 0.83 & 0.84 \\
 \hline
 COCO & YT-BB & no & 0.77 & 0.78 \\
 COCO & YT-BB & yes & 0.77 & 0.78 \\
 \hline
 YT-BB & YT-BB & no & 0.93 & 0.94 \\
 YT-BB & YT-BB & yes & 0.95 & 0.96 \\
 \hline
 YT-BB & COCO & - & 0.66 & 0.67 \\
 \hline
\end{tabular}
\end{center}
\caption{Summary of image classification baselines. mAP and AUC are calculated across the 23 object classes (excluding ``NONE''). The ``smooth?'' column indicates whether the predictions of YT-BB were averaged in time.}
\label{table:baseline-classification}
\end{table}

We started by comparing the relative difficulty of two instances of the same image classification model, one trained on YT-BB (``the YT-BB model'') and one trained on COCO (``the COCO model''). Our data set has explicit classification annotations. For COCO, we treated the presence or absence of any object localization of a class as either a positive or negative label, respectively. Both models employed an Inception-v3\footnote{\label{github_footnote} See Supplementary Section~\ref{github_section} for GitHub locations.} architecture \cite{szegedy2015} with logistic regression, implemented in TensorFlow\cite{tensorflow2015-whitepaper}. The choice of logistic regression reflects the fact that multiple labels may be associated with a single image. Both models were initialized with the weights of an Inception-v3 image classification system pre-trained on the ImageNet 2012 Challenge data set \cite{deng2009imagenet} and subsequently fine-tuned on YT-BB/COCO individually.

We measured the mean precision-recall curve across all 23 classes (excluding the ``NONE'' class since it is not available in COCO).  These results are shown in the dark curves in Figure \ref{fig:baseline-classification}. We find that training on YT-BB (mAP = 0.93) is easier than on COCO (mAP = 0.83), which could reflect the larger amount of training data per-class available in YT-BB.

One open question is the difficulty of domain transfer--i.e. training on one data set and evaluating on the other. We assessed this by measuring the mean precision-recall curve across 23 classes for the COCO model on YT-BB data (Figure \ref{fig:baseline-classification}, right panel, light curve) and vice-versa (Figure \ref{fig:baseline-classification}, left panel, light curve).  We find that a COCO model evaluated on YT-BB (mAP = 0.77) was worse than one evaluated on COCO data (mAP = 0.83). The analogous claim is true for a model trained on YT-BB (Table \ref{table:baseline-classification}). These results indicate that images in YT-BB are diverse and not just a subset of those in COCO.

%%%%%%%%%%%%%%%%%%%%%%%%%%%%%%%%%%%%%%%%%%%%%%%%%%%%%%%%%%%%%%%%%%%%%%%%%%%%%%%%%%%%%%%%%%%%%%%%%%%
%%%%%%%%% MULTI-BOX BASELINE %%%%%%%%%%%%%%%%%%%%%%%%%%%%%%%%%%%%%%%%%%%%%%%%%%%%%%%%%%%%%%%%%%%%%%
%%%%%%%%%%%%%%%%%%%%%%%%%%%%%%%%%%%%%%%%%%%%%%%%%%%%%%%%%%%%%%%%%%%%%%%%%%%%%%%%%%%%%%%%%%%%%%%%%%%
\subsection{Object detection}

\begin{figure}[t]
\begin{center}
   \includegraphics[width=0.95\linewidth]{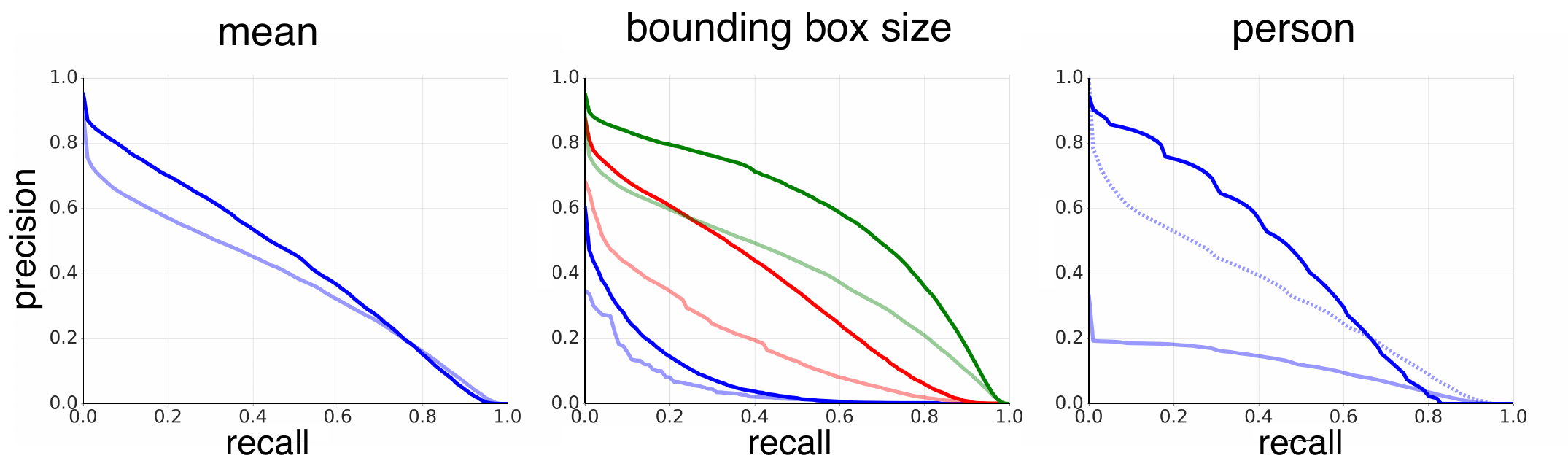}
\end{center}
   \caption{Baseline comparison of the COCO detection model.
   Precision-recall curve when evaluated on COCO~\cite{lin2014microsoft} (dark line) and on YT-BB (light line). Left: Mean across the 23 classes. Center: Delineated by the bounding box size into large (green; top 2 curves), medium (red; middle 2 curves) and small (blue; bottom 2 curves). Right: Detection of the person class on COCO (dark top line), on YT-BB (light bottom line), and on person-specific subset of YT-BB (dashed middle line)}
\label{fig:baseline-BoundingBoxes}
\end{figure}

\begin{table*}
\begin{center}
\begin{tabular}{ |c|c|c|c|c|c|c|c| } 
 \hline
 Training data & Evaluation data & mAP & mAP @50\% & mAP @75\% & mAP small & mAP medium & mAP large \\
 \hline
 COCO* & COCO* & 0.33 & 0.54 & 0.34 & 0.06 & 0.29 & 0.49 \\
 \hline
 COCO & COCO & 0.43 & 0.67 & 0.47 & 0.08 & 0.35 & 0.58 \\
 COCO & YT-BB & 0.37 & 0.56 & 0.41 & 0.05 & 0.18 & 0.41 \\
 YT-BB & YT-BB & 0.59 & 0.81 & 0.66 & 0.07 & 0.31 & 0.63 \\
 YT-BB & COCO & 0.31 & 0.47 & 0.34 & 0.01 & 0.16 & 0.46 \\
 \hline
\end{tabular}
\end{center}
\caption{Summary of image detection baselines between COCO and YT-BB across the 23 label classes. mAP is the mean average precision, averaged over multiple categories, multiple scales and multiple IOU fractions ranging between 0.5 and
0.95. mAP is the COCO competition metric. mAP @50 is the same but restricted to IOU $\geq$ 50\%. mAP @75 is the
same but restricted to IOU $\geq$ 75\%. mAP (size) is the same but restricted to small, medium and large objects. For reference, the top row (COCO*) highlights a single-crop, no-ensemble version of the winning entry to the COCO competition as measured across all 80 COCO classes.}
\label{table:baseline-BoundingBoxes}
\end{table*}

\begin{figure}[t]
\begin{center}
   \includegraphics[width=0.95\linewidth]{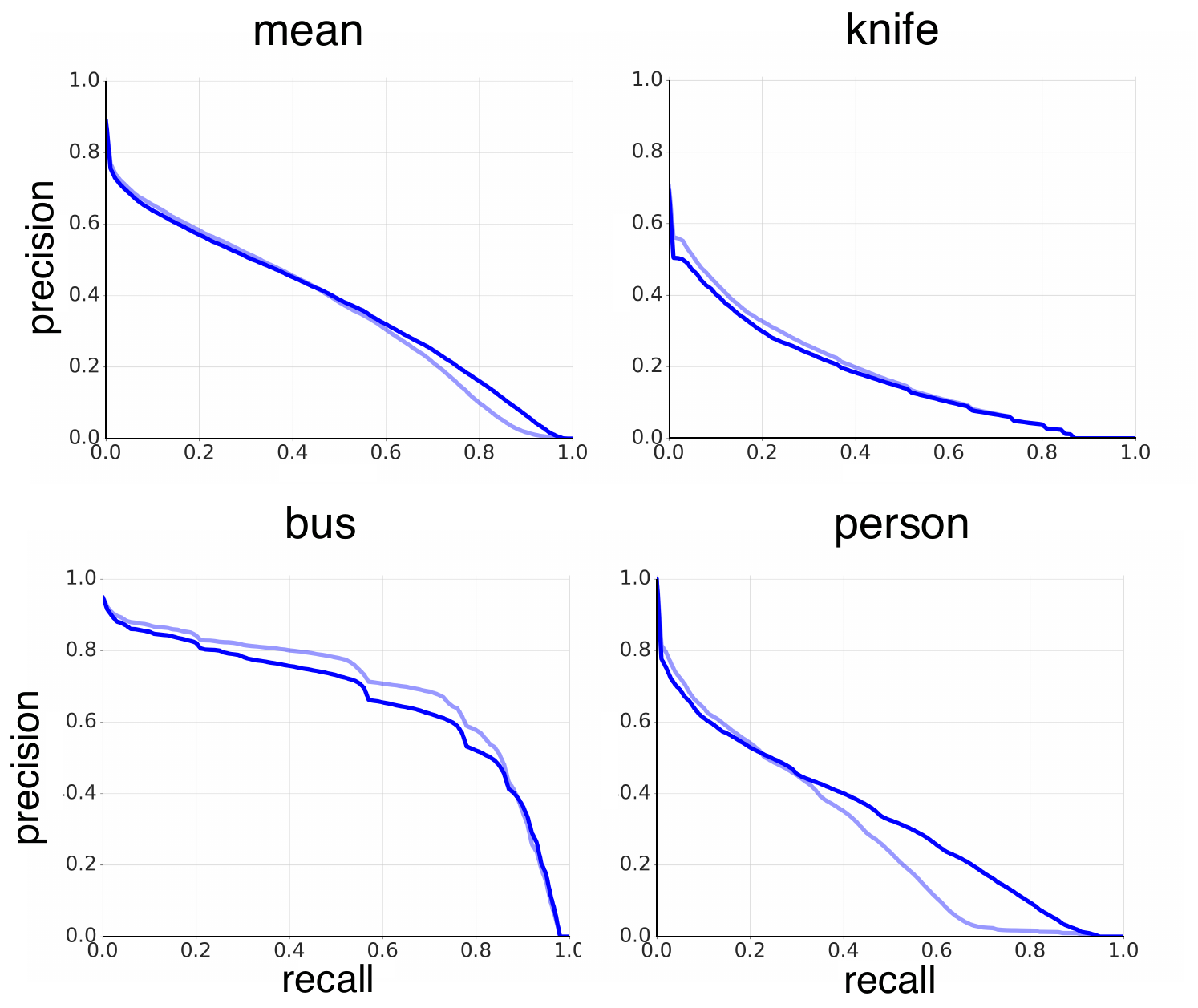}
\end{center}
   \caption{Effect of simple temporal smoothing on localization baselines. The dark curve shows precision-recall curve for single-frame object localization
(mAP = 0.37, 0.18, 0.66, 0.33) and the light curve shows precision-recall employing simple temporal smoothing (mAP = 0.36, 0.19, 0.69, 0.28, respectively). See text for details.}
\label{fig:localization-smoothing}
\end{figure}

We then compared the relative difficulty of YT-BB and COCO for object detection. We used two instances of a Faster-RCNN\textsuperscript{\ref{github_footnote}} detection proposal architecture \cite{huang2016speed} paired with an Inception-ResNet-v2\textsuperscript{\ref{github_footnote}} feature network \cite{ren2015faster, szegedy2016}. Increasing the number of detection proposal results in improved object localizations at the expense of more computationally expensive inference and training. We selected a set of hyperparameters that resulted in 1400b FLOPs per frame for inference. Both instances were partially initialized with the weights of an Inception-ResNet-v2 image classification system trained on the ImageNet 2012 Challenge data set \cite{deng2009imagenet}. One instance was subsequently trained further on YT-BB (``the YT-BB model'') and another on COCO (``the COCO model'').

We evaluated the performance of each model by measuring the mean precision-recall curve across all 23 classes. These results are summarized across several
standard calculations of mAP for object detection in Table~\ref{table:baseline-BoundingBoxes} and delineated by class in Supplementary Table \ref{tab:perclass}. We find that training on YT-BB (mAP = 0.59) is easier than on COCO (mAP = 0.43). This result is consistent when measured across a range of detection box sizes.

In parallel with the image classification baseline, we also measured the relative difficulty of the two data sets by considering the problem of domain transfer. Again, we assessed this by examining the mean precision-recall curve across the 23 classes for the COCO model evaluated on YT-BB data.
The degree to which the COCO model performance on YT-BB was worse then COCO reflects the relative difficulty and domain shift of the YT-BB data set.
We found that a COCO model evaluated on YT-BB (mAP = 0.37) was indeed worse than when evaluated on COCO data (mAP = 0.43). This result was consistent across bounding box sizes and ranges of overlap assessment (Table \ref{table:baseline-BoundingBoxes}, Figure \ref{fig:baseline-BoundingBoxes}). Notably, the COCO model was particularly poor at localizing medium and large YT-BB objects (Figure \ref{fig:baseline-BoundingBoxes}, middle panel). The analogous claim is true for a model trained on YT-BB (Table \ref{table:baseline-BoundingBoxes}).

We next focused on the ``person'' class. The COCO model performed significantly worse in this case (mAP = 0.41 vs mAP = 0.12).  At the lowest possible threshold, the COCO model fails to identify more than $\sim 62\%$ of the ``person'' detections in YT-BB frames (Figure \ref{fig:baseline-BoundingBoxes}, right panel, light curve). At high thresholds, the COCO model exhibits low precision for ``person''. This may be due to the fact that YT-BB is not exhaustively labeled. Unlabeled people may appear in videos which have been annotated for other classes. This may result in high false-positive scores and systematically lower precision. To mitigate this artifact we restrict the evaluation of the COCO model to a subset of YT-BB frames that have been labeled with a bounding box for ``person'' (Figure \ref{fig:baseline-BoundingBoxes}, right panel, dashed curve). The precision-recall curve was lifted as a result of the removal of many unlabeled people, but remained below the precision-recall curve evaluated on COCO data. This analysis is however imperfect since images annotated with a ``person'' localization might contain additional people. Future work will be needed to determine how much of the additional difficulty ascribed to the YT-BB ``person'' label is due to the diversity of the ``person'' poses available in the YT-BB data set.

\subsection{Exploiting temporal information in videos}

All our baselines up to this point treated the frames as individual images. A salient aspect of the YT-BB data set is, however, that these frames exist within contiguous segments of video. Such video sequence can help regularize and improve video frame predictions. Devising better learning architectures for this purpose is an area of intense research interest \cite{shortsnippets, ranzato, simonyan}. As a demonstration of this data set's potential, we performed several simple manipulations that indicate that temporal information exists and may be used by a learning system.

For the image classification task, we replaced the prediction for each label with the mean prediction for each label across each YT-BB video segment. The result of this temporal smoothing is shown in the dashed red line in Figure \ref{fig:baseline-classification} and summarized in Table \ref{table:baseline-classification}. Although the mAP and AUC do not change significantly (Table \ref{table:baseline-classification}), the precision-recall curves do highlight that the temporally-averaged prediction systematically surpasses the single-frame prediction in the high recall regime (\eg recall $>\, \sim 0.7$). In principle, one could therefore build an improved system which achieves the \textit{envelope} of the single-frame and temporally-averaged prediction scores.

For the object detection task, we down-weighted spurious weak object detections that appeared in single video frames but not in neighboring frames. Specifically, we artificially multiplied by $\frac{1}{100}$ the confidence scores of detected objects that did not overlap significantly with previous and subsequent frame detections (IOU $<\, 0.1$; confidence $<\, 0.5$). Figure \ref{fig:localization-smoothing} shows the effects of this manipulation on the precision-recall curves. When aggregating across all classes, temporal smoothing slightly reduces model performance (mAP = 0.36 vs mAP = 0.37). We broke down this result to expose the diversity of behavior across labels. This revealed elevated precision-recall curves for some classes (\eg ``knife'', ``bus'') but lowered curves for other classes (\eg ``person''). In principle, a unified model could at least learn which classes benefit and apply the correction only to those. Despite the mixed results, this analysis suggests that taking into account the temporal structure of the video could result in better detection models.

%%%%%%%%%%%%%%%%%%%%%%%%%%%%%%%%%%%%%%%%%%%%%%%%%%%%%%%%%%%%%%%%%%%%%%%%%%%%%%%%%%%%%%%%%%%%%%%%%%%
%%%%%%%%%%%%%%%%%%%%%%%%%%%%%%%%%%%%%%%%%%%%%%%%%%%%%%%%%%%%%%%%%%%%%%%%%%%%%%%%%%%%%%%%%%%%%%%%%%%
%%%%%%%%% DISCUSSION %%%%%%%%%%%%%%%%%%%%%%%%%%%%%%%%%%%%%%%%%%%%%%%%%%%%%%%%%%%%%%%%%%%%%%%%%%%%%%
%%%%%%%%%%%%%%%%%%%%%%%%%%%%%%%%%%%%%%%%%%%%%%%%%%%%%%%%%%%%%%%%%%%%%%%%%%%%%%%%%%%%%%%%%%%%%%%%%%%
%%%%%%%%%%%%%%%%%%%%%%%%%%%%%%%%%%%%%%%%%%%%%%%%%%%%%%%%%%%%%%%%%%%%%%%%%%%%%%%%%%%%%%%%%%%%%%%%%%%
\section{Discussion}
\label{sec:discussion}

In this paper, we introduced YT-BB, a new data set with 380,000 video segments, annotated with 5.6 million human-drawn bounding boxes tracking everyday objects in 23 categories. This represents an unprecedentedly large video detection data set (Section \ref{sec:related_work}). First, we described the data mining process that led to minimally-edited videos and the human annotation stages that produced tight and precise bounding boxes, as well as precise tags indicating object absence (Sections \ref{sec:data_mining} and \ref{sec:human_annotation}). Then we presented relevant statistics and measures of annotation quality for each class. In particular, basic metrics of bounding box motion indicate that the objects or the camera exhibit significant changes throughout the segment (Section \ref{sec:results}). Finally, we showed baselines for classification and detection trained and evaluated both on this data set and on COCO. These baselines demonstrate the potential for information in the video sequence to improve upon the basic inferences that can be done from single frames alone (Section \ref{sec:baseline_models}).

Future work could refine YT-BB in various ways, most notably by adding more classes. With only 23 classes, we were able to pay special attention to each (Supplementary Section \ref{sec:attention_span}). Scaling up, classes would have to be treated more generically, as was done in \cite{deng2009imagenet}. This would magnify the challenges of crowd-sourcing schemes (``paradigm A''), in which low annotator accountability produces initial answers that often have poor quality \cite{ipeirotis2010quality}, requiring significant additional effort to get to the final labels \cite{paritosh2012human, sheng2008get, waterhouse2013pay}. In this work, we observed such challenges in stages 1 and 2 (Section \ref{sec:human_annotation} and Supplementary Section \ref{sec:attention_span}). Alternatively, one could use a group of dedicated annotators who are committed to the project (``paradigm B''), as we did in stages 3 and 4 (Section \ref{sec:human_annotation}). While we never carried out a proper A/B test, anecdotally we found paradigm B much more satisfying for a large-scale project. This can be traced back to the ability to train the annotators \cite{dow2011shepherding} and to provide them with feedback over time, resolving each problem encountered ``once-and-for-all''.

Another direction for improvement could be to gather more bounding boxes. Increasing the sheer number does not seem critical as our baselines show no signs of over-fitting. On the other hand, exhaustively labeling the existing videos may prove helpful, especially within the testing subset. While this would render the annotation task more complex, simplicity could be regained by introducing additional stages. Cascading stages have been found useful before \cite{lin2014microsoft}. In our case, it allowed the tuning of the annotation tool's user interface to each task (Supplementary Section \ref{sec:human_annotation_user_interfaces}), rendering the first stage as much as $50$ times faster than the last one. This, in turn, allowed for more negative examples to be present at the input since they could be easily discarded, and therefore the initial data mining stage could be more permissive. User interface optimization sometimes yielded unexpected results. For example, it turned out that providing default guesses for the bounding box locations was often {\it not} faster. Moreover, the annotators may find it easier to leave the default unchanged, which could bias the results toward such automatically generated defaults. Removing these defaults also made the tool simpler, which is generally known to be advantageous \cite{finnerty2013keep}.

The baseline results suggest that there exists headroom for improving the quality of models on this data set. In particular, the data affords two distinct research directions. One is that the human annotation results identified individual video frames that are \textit{hard negatives}, i.e. individual frames in the video that did not contain the object of interest even though surrounding frames did. These hard negatives might provide useful training and evaluation examples for future visual models.

The second research direction is to build models that harness the information in the temporal sequence of frames in a computationally efficient manner. Our baseline results indicate that even performing naive manipulations that incorporate such temporal aspects may contribute to better object classification and detection in video. The ability to build tractable, scalable models that exploit sequential information by keeping an internal memory state (\eg \cite{gru, lstm}) would likely lead toward better object detection and tracking (\eg \cite{deeptrack, wang2015visual}).\footnote{The data is available at https://research.google.com/youtube-bb}

%%%%%%%%%%%%%%%%%%%%%%%%%%%%%%%%%%%%%%%%%%%%%%%%%%%%%%%%%%%%%%%%%%%%%%%%%%%%%%%%%%%%%%%%%%%%%%%%%%%
%%%%%%%%%%%%%%%%%%%%%%%%%%%%%%%%%%%%%%%%%%%%%%%%%%%%%%%%%%%%%%%%%%%%%%%%%%%%%%%%%%%%%%%%%%%%%%%%%%%
%%%%%%%%% ACKNOWLEDGEMENTS %%%%%%%%%%%%%%%%%%%%%%%%%%%%%%%%%%%%%%%%%%%%%%%%%%%%%%%%%%%%%%%%%%%%%%%%
%%%%%%%%%%%%%%%%%%%%%%%%%%%%%%%%%%%%%%%%%%%%%%%%%%%%%%%%%%%%%%%%%%%%%%%%%%%%%%%%%%%%%%%%%%%%%%%%%%%
%%%%%%%%%%%%%%%%%%%%%%%%%%%%%%%%%%%%%%%%%%%%%%%%%%%%%%%%%%%%%%%%%%%%%%%%%%%%%%%%%%%%%%%%%%%%%%%%%%%
\section{Acknowledgements}
We wish to thank Matthias Grundmann, John Gregg, Christian Falk and especially Thomas Silva for early efforts at bounding box annotations; Susanna Ricco, Sanketh Shetty for general advice about mining video data; George Toderici, Rahul Sukthankar for advice in many aspects of this work, Sami Abu-El-Haija, Manfred Georg for enormous efforts and generous advice about harvesting and annotating YouTube videos; Mir Shabber Ali Khan, Ashwin Kakarla and many others for the human annotations; and the larger Google Brain team for support with TensorFlow and training vision models.

%%%%%%%%%%%%%%%%%%%%%%%%%%%%%%%%%%%%%%%%%%%%%%%%%%%%%%%%%%%%%%%%%%%%%%%%%%%%%%%%%%%%%%%%%%%%%%%%%%%
%%%%%%%%%%%%%%%%%%%%%%%%%%%%%%%%%%%%%%%%%%%%%%%%%%%%%%%%%%%%%%%%%%%%%%%%%%%%%%%%%%%%%%%%%%%%%%%%%%%
%%%%%%%%% BIBLIOGRAPHY %%%%%%%%%%%%%%%%%%%%%%%%%%%%%%%%%%%%%%%%%%%%%%%%%%%%%%%%%%%%%%%%%%%%%%%%%%%%
%%%%%%%%%%%%%%%%%%%%%%%%%%%%%%%%%%%%%%%%%%%%%%%%%%%%%%%%%%%%%%%%%%%%%%%%%%%%%%%%%%%%%%%%%%%%%%%%%%%
%%%%%%%%%%%%%%%%%%%%%%%%%%%%%%%%%%%%%%%%%%%%%%%%%%%%%%%%%%%%%%%%%%%%%%%%%%%%%%%%%%%%%%%%%%%%%%%%%%%

{\small
\bibliographystyle{ieee}
\bibliography{yt_bb}

\begin{thebibliography}{10}\itemsep=-1pt

\bibitem{tensorflow2015-whitepaper}
M.~Abadi, A.~Agarwal, P.~Barham, E.~Brevdo, Z.~Chen, C.~Citro, G.~S. Corrado,
  A.~Davis, J.~Dean, M.~Devin, S.~Ghemawat, I.~Goodfellow, A.~Harp, G.~Irving,
  M.~Isard, Y.~Jia, R.~Jozefowicz, L.~Kaiser, M.~Kudlur, J.~Levenberg,
  D.~Man\'{e}, R.~Monga, S.~Moore, D.~Murray, C.~Olah, M.~Schuster, J.~Shlens,
  B.~Steiner, I.~Sutskever, K.~Talwar, P.~Tucker, V.~Vanhoucke, V.~Vasudevan,
  F.~Vi\'{e}gas, O.~Vinyals, P.~Warden, M.~Wattenberg, M.~Wicke, Y.~Yu, and
  X.~Zheng.
\newblock {TensorFlow}: Large-scale machine learning on heterogeneous systems,
  2015.
\newblock Software available from tensorflow.org.

\bibitem{1609.08675}
S.~Abu-El-Haija, N.~Kothari, J.~Lee, P.~Natsev, G.~Toderici, B.~Varadarajan,
  and S.~Vijayanarasimhan.
\newblock Youtube-8m: A large-scale video classification benchmark, 2016.

\bibitem{2016trecvidawad}
G.~Awad, J.~Fiscus, M.~Michel, D.~Joy, W.~Kraaij, A.~F. Smeaton, G.~Quénot,
  M.~Eskevich, R.~Aly, G.~J.~F. Jones, R.~Ordelman, B.~Huet, and M.~Larson.
\newblock Trecvid 2016: Evaluating video search, video event detection,
  localization, and hyperlinking.
\newblock In {\em Proceedings of TRECVID 2016}. NIST, USA, 2016.

\bibitem{bernstein2015soylent}
M.~S. Bernstein, G.~Little, R.~C. Miller, B.~Hartmann, M.~S. Ackerman, D.~R.
  Karger, D.~Crowell, and K.~Panovich.
\newblock Soylent: a word processor with a crowd inside.
\newblock {\em Communications of the ACM}, 58(8):85--94, 2015.

\bibitem{buhrmester2011amazon}
M.~Buhrmester, T.~Kwang, and S.~D. Gosling.
\newblock Amazon's mechanical turk a new source of inexpensive, yet
  high-quality, data?
\newblock {\em Perspectives on psychological science}, 6(1):3--5, 2011.

\bibitem{caffe_model_zoo}
{Caffe Model Zoo}.
\newblock http://github.com/BVLC/caffe/wiki/Model-Zoo.
\newblock [Accessed 19-Oct-2016].

\bibitem{chen2011opportunities}
J.~J. Chen, N.~J. Menezes, A.~D. Bradley, and T.~North.
\newblock Opportunities for crowdsourcing research on amazon mechanical turk.
\newblock {\em Interfaces}, 5(3), 2011.

\bibitem{gru}
K.~Cho, B.~van Merrienboer, {\c{C}}.~G{\"{u}}l{\c{c}}ehre, F.~Bougares,
  H.~Schwenk, and Y.~Bengio.
\newblock Learning phrase representations using {RNN} encoder-decoder for
  statistical machine translation.
\newblock {\em CoRR}, abs/1406.1078, 2014.

\bibitem{deng2009imagenet}
J.~Deng, W.~Dong, R.~Socher, L.-J. Li, K.~Li, and L.~Fei-Fei.
\newblock Imagenet: A large-scale hierarchical image database.
\newblock In {\em Computer Vision and Pattern Recognition, 2009. CVPR 2009.
  IEEE Conference on}, pages 248--255. IEEE, 2009.

\bibitem{dollar2009pedestrian}
P.~Doll{\'a}r, C.~Wojek, B.~Schiele, and P.~Perona.
\newblock Pedestrian detection: A benchmark.
\newblock In {\em Computer Vision and Pattern Recognition, 2009. CVPR 2009.
  IEEE Conference on}, pages 304--311. IEEE, 2009.

\bibitem{dow2011shepherding}
S.~Dow, A.~Kulkarni, B.~Bunge, T.~Nguyen, S.~Klemmer, and B.~Hartmann.
\newblock Shepherding the crowd: managing and providing feedback to crowd
  workers.
\newblock In {\em CHI'11 Extended Abstracts on Human Factors in Computing
  Systems}, pages 1669--1674. ACM, 2011.

\bibitem{everingham2015pascal}
M.~Everingham, S.~A. Eslami, L.~Van~Gool, C.~K. Williams, J.~Winn, and
  A.~Zisserman.
\newblock The pascal visual object classes challenge: A retrospective.
\newblock {\em International Journal of Computer Vision}, 111(1):98--136, 2015.

\bibitem{everingham2010pascal}
M.~Everingham, L.~Van~Gool, C.~K. Williams, J.~Winn, and A.~Zisserman.
\newblock The pascal visual object classes (voc) challenge.
\newblock {\em International journal of computer vision}, 88(2):303--338, 2010.

\bibitem{fei2007learning}
L.~Fei-Fei, R.~Fergus, and P.~Perona.
\newblock Learning generative visual models from few training examples: An
  incremental bayesian approach tested on 101 object categories.
\newblock {\em Computer Vision and Image Understanding}, 106(1):59--70, 2007.

\bibitem{wordnet}
C.~Fellbaum.
\newblock {\em WordNet: An Electronic Lexical Database}.
\newblock Bradford Books, 1998.

\bibitem{finnerty2013keep}
A.~Finnerty, P.~Kucherbaev, S.~Tranquillini, and G.~Convertino.
\newblock Keep it simple: Reward and task design in crowdsourcing.
\newblock In {\em Proceedings of the Biannual Conference of the Italian Chapter
  of SIGCHI}, page~14. ACM, 2013.

\bibitem{THUMOS15}
A.~Gorban, H.~Idrees, Y.-G. Jiang, A.~Roshan~Zamir, I.~Laptev, M.~Shah, and
  R.~Sukthankar.
\newblock {THUMOS} challenge: Action recognition with a large number of
  classes.
\newblock http://www.thumos.info, 2015.

\bibitem{pamir}
D.~Grangier and S.~Bengio.
\newblock A discriminative kernel-based model to rank images from text queries.
\newblock {\em IEEE Transactions on Pattern Analysis and Machine Intelligence
  (TPAMI)}, 2008.

\bibitem{griffin2007caltech}
G.~Griffin, A.~Holub, and P.~Perona.
\newblock Caltech-256 object category dataset.
\newblock 2007.

\bibitem{he2015deep}
K.~He, X.~Zhang, S.~Ren, and J.~Sun.
\newblock Deep residual learning for image recognition.
\newblock {\em arXiv preprint arXiv:1512.03385}, 2015.

\bibitem{lstm}
S.~Hochreiter and J.~Schmidhuber.
\newblock Long short-term memory.
\newblock {\em Neural Comput.}, 9(8):1735--1780, Nov. 1997.

\bibitem{huang2016speed}
J.~Huang, V.~Rathod, C.~Sun, M.~Zhu, A.~Korattikara, A.~Fathi, I.~Fischer,
  Z.~Wojna, Y.~Song, S.~Guadarrama, et~al.
\newblock Speed/accuracy trade-offs for modern convolutional object detectors.
\newblock {\em arXiv preprint arXiv:1611.10012}, 2016.

\bibitem{ipeirotis2010quality}
P.~G. Ipeirotis, F.~Provost, and J.~Wang.
\newblock Quality management on amazon mechanical turk.
\newblock In {\em Proceedings of the ACM SIGKDD workshop on human computation},
  pages 64--67. ACM, 2010.

\bibitem{caffe}
Y.~Jia, E.~Shelhamer, J.~Donahue, S.~Karayev, J.~Long, R.~Girshick,
  S.~Guadarrama, and T.~Darrell.
\newblock Caffe: Convolutional architecture for fast feature embedding.
\newblock In {\em Proceedings of the 22Nd ACM International Conference on
  Multimedia}, MM '14, pages 675--678, New York, NY, USA, 2014. ACM.

\bibitem{KarpathyCVPR14}
A.~Karpathy, G.~Toderici, S.~Shetty, T.~Leung, R.~Sukthankar, and L.~Fei-Fei.
\newblock Large-scale video classification with convolutional neural networks.
\newblock In {\em CVPR}, 2014.

\bibitem{youtube_sports}
A.~Karpathy, G.~Toderici, S.~Shetty, T.~Leung, R.~Sukthankar, and L.~Fei-Fei.
\newblock {The YouTube Sports-1M Dataset}.
\newblock http://github.com/gtoderici/sports-1m-dataset, 2014.
\newblock [Accessed 19-Oct-2016].

\bibitem{kristan2015visual}
M.~Kristan, J.~Matas, A.~Leonardis, M.~Felsberg, L.~Cehovin, G.~Fernandez,
  T.~Vojir, G.~Hager, G.~Nebehay, and R.~Pflugfelder.
\newblock The visual object tracking vot2015 challenge results.
\newblock In {\em Proceedings of the IEEE International Conference on Computer
  Vision Workshops}, pages 1--23, 2015.

\bibitem{krizhevsky2012imagenet}
A.~Krizhevsky, I.~Sutskever, and G.~E. Hinton.
\newblock Imagenet classification with deep convolutional neural networks.
\newblock In {\em Advances in neural information processing systems}, pages
  1097--1105, 2012.

\bibitem{Kuehne11}
H.~Kuehne, H.~Jhuang, E.~Garrote, T.~Poggio, and T.~Serre.
\newblock {HMDB}: a large video database for human motion recognition.
\newblock In {\em Proceedings of the International Conference on Computer
  Vision (ICCV)}, 2011.

\bibitem{MOTChallenge2015}
L.~Leal-Taix\'{e}, A.~Milan, I.~Reid, S.~Roth, and K.~Schindler.
\newblock {MOTC}hallenge 2015: {T}owards a benchmark for multi-target tracking.
\newblock {\em arXiv:1504.01942 [cs]}, Apr. 2015.
\newblock arXiv: 1504.01942.

\bibitem{deeptrack}
H.~Li, Y.~Li, and F.~Porikli.
\newblock Deeptrack: Learning discriminative feature representations online for
  robust visual tracking.
\newblock {\em CoRR}, abs/1503.00072, 2015.

\bibitem{lin2014microsoft}
T.-Y. Lin, M.~Maire, S.~Belongie, J.~Hays, P.~Perona, D.~Ramanan,
  P.~Doll{\'a}r, and C.~L. Zitnick.
\newblock Microsoft coco: Common objects in context.
\newblock In {\em European Conference on Computer Vision}, pages 740--755.
  Springer, 2014.

\bibitem{martin2001database}
D.~Martin, C.~Fowlkes, D.~Tal, and J.~Malik.
\newblock A database of human segmented natural images and its application to
  evaluating segmentation algorithms and measuring ecological statistics.
\newblock In {\em Computer Vision, 2001. ICCV 2001. Proceedings. Eighth IEEE
  International Conference on}, volume~2, pages 416--423. IEEE, 2001.

\bibitem{nene1996columbia}
S.~A. Nene, S.~K. Nayar, H.~Murase, et~al.
\newblock Columbia object image library (coil-20).
\newblock Technical report, Technical report CUCS-005-96, 1996.

\bibitem{shortsnippets}
J.~Y.-H. Ng, M.~Hausknecht, S.~Vijayanarasimhan, O.~Vinyals, R.~Monga, and
  G.~Toderici.
\newblock Beyond short snippets: Deep networks for video classification.
\newblock In {\em Computer Vision and Pattern Recognition}, 2015.

\bibitem{conse}
M.~Norouzi, T.~Mikolov, S.~Bengio, Y.~Singer, J.~Shlens, A.~Frome, G.~Corrado,
  and J.~Dean.
\newblock Zero-shot learning by convex combination of semantic embeddings.
\newblock {\em CoRR}, abs/1312.5650, 2013.

\bibitem{paritosh2012human}
P.~Paritosh.
\newblock Human computation must be reproducible.
\newblock 2012.

\bibitem{prest2012learning}
A.~Prest, C.~Leistner, J.~Civera, C.~Schmid, and V.~Ferrari.
\newblock Learning object class detectors from weakly annotated video.
\newblock In {\em Computer Vision and Pattern Recognition (CVPR), 2012 IEEE
  Conference on}, pages 3282--3289. IEEE, 2012.

\bibitem{youtube_objects}
A.~Prest, C.~Leistner, J.~Civera, C.~Schmid, and V.~Ferrari.
\newblock {Youtube-Objects dataset}.
\newblock https://data.vision.ee.ethz.ch/cvl/youtube-objects/, 2012.
\newblock [Accessed 19-Oct-2016].

\bibitem{ranzato}
M.~Ranzato, A.~Szlam, J.~Bruna, M.~Mathieu, R.~Collobert, and S.~Chopra.
\newblock Video (language) modeling: a baseline for generative models of
  natural videos.
\newblock {\em CoRR}, abs/1412.6604, 2014.

\bibitem{ren2015faster}
S.~Ren, K.~He, R.~Girshick, and J.~Sun.
\newblock Faster r-cnn: Towards real-time object detection with region proposal
  networks.
\newblock In {\em Advances in neural information processing systems}, pages
  91--99, 2015.

\bibitem{russakovsky2013detecting}
O.~Russakovsky, J.~Deng, Z.~Huang, A.~C. Berg, and L.~Fei-Fei.
\newblock Detecting avocados to zucchinis: what have we done, and where are we
  going?
\newblock In {\em Proceedings of the IEEE International Conference on Computer
  Vision}, pages 2064--2071, 2013.

\bibitem{ILSVRC15}
O.~Russakovsky, J.~Deng, H.~Su, J.~Krause, S.~Satheesh, S.~Ma, Z.~Huang,
  A.~Karpathy, A.~Khosla, M.~Bernstein, A.~C. Berg, and L.~Fei-Fei.
\newblock {ImageNet Large Scale Visual Recognition Challenge}.
\newblock {\em International Journal of Computer Vision (IJCV)},
  115(3):211--252, 2015.

\bibitem{sheng2008get}
V.~Sheng, F.~Provost, and P.~Ipeirotis.
\newblock Get another label? improving data quality and data mining.
\newblock 2008.

\bibitem{simonyan}
K.~Simonyan and A.~Zisserman.
\newblock Two-stream convolutional networks for action recognition in videos.
\newblock {\em CoRR}, abs/1406.2199, 2014.

\bibitem{simonyan_vgg}
K.~Simonyan and A.~Zisserman.
\newblock Very deep convolutional networks for large-scale image recognition.
\newblock {\em CoRR}, abs/1409.1556, 2014.

\bibitem{soomro2012ucf101}
K.~Soomro, A.~R. Zamir, and M.~Shah.
\newblock Ucf101: A dataset of 101 human actions classes from videos in the
  wild.
\newblock {\em arXiv preprint arXiv:1212.0402}, 2012.

\bibitem{szegedy2016}
C.~Szegedy, S.~Ioffe, and V.~Vanhoucke.
\newblock Inception-v4, inception-resnet and the impact of residual connections
  on learning.
\newblock {\em arXiv preprint arXiv:1602.07261}, 2016.

\bibitem{szegedy2015}
C.~Szegedy, V.~Vanhoucke, S.~Ioffe, J.~Shlens, and Z.~Wojna.
\newblock Rethinking the inception architecture for computer vision.
\newblock {\em arXiv preprint arXiv:1512.00567}, 2015.

\bibitem{couchpotato}
K.~Tang, R.~Sukthankar, J.~Yagnik, and L.~Fei-Fei.
\newblock Discriminative segment annotation in weakly labeled video.
\newblock In {\em Proceedings of International Conference on Computer Vision
  and Pattern Recognition (CVPR 2013)}, 2013.

\bibitem{tfslim}
{TensorFlow-Slim image classification library}.
\newblock http://github.com/tensorflow/models/tree/master/slim.
\newblock [Accessed 19-Oct-2016].

\bibitem{torralba2004sharing}
A.~Torralba, K.~P. Murphy, and W.~T. Freeman.
\newblock Sharing features: efficient boosting procedures for multiclass object
  detection.
\newblock In {\em Computer Vision and Pattern Recognition, 2004. CVPR 2004.
  Proceedings of the 2004 IEEE Computer Society Conference on}, volume~2, pages
  II--762. IEEE, 2004.

\bibitem{caption}
O.~Vinyals, A.~Toshev, S.~Bengio, and D.~Erhan.
\newblock Show and tell: {A} neural image caption generator.
\newblock {\em CoRR}, abs/1411.4555, 2014.

\bibitem{wang2015visual}
L.~Wang, W.~Ouyang, X.~Wang, and H.~Lu.
\newblock Visual tracking with fully convolutional networks.
\newblock In {\em IEEE International Conference on Computer Vision (ICCV)},
  2015.

\bibitem{waterhouse2013pay}
T.~P. Waterhouse.
\newblock Pay by the bit: an information-theoretic metric for collective human
  judgment.
\newblock In {\em Proceedings of the 2013 conference on Computer supported
  cooperative work}, pages 623--638. ACM, 2013.

\bibitem{xiao2014sun}
J.~Xiao, K.~A. Ehinger, J.~Hays, A.~Torralba, and A.~Oliva.
\newblock Sun database: Exploring a large collection of scene categories.
\newblock {\em International Journal of Computer Vision}, pages 1--20, 2014.

\bibitem{xiao2010sun}
J.~Xiao, J.~Hays, K.~A. Ehinger, A.~Oliva, and A.~Torralba.
\newblock Sun database: Large-scale scene recognition from abbey to zoo.
\newblock In {\em Computer vision and pattern recognition (CVPR), 2010 IEEE
  conference on}, pages 3485--3492. IEEE, 2010.

\end{thebibliography}
}

%%%%%%%%%%%%%%%%%%%%%%%%%%%%%%%%%%%%%%%%%%%%%%%%%%%%%%%%%%%%%%%%%%%%%%%%%%%%%%%%%%%%%%%%%%%%%%%%%%%
%%%%%%%%%%%%%%%%%%%%%%%%%%%%%%%%%%%%%%%%%%%%%%%%%%%%%%%%%%%%%%%%%%%%%%%%%%%%%%%%%%%%%%%%%%%%%%%%%%%
%%%%%%%%% SUPPLEMENTARY MATERIALS %%%%%%%%%%%%%%%%%%%%%%%%%%%%%%%%%%%%%%%%%%%%%%%%%%%%%%%%%%%%%%%%%
%%%%%%%%%%%%%%%%%%%%%%%%%%%%%%%%%%%%%%%%%%%%%%%%%%%%%%%%%%%%%%%%%%%%%%%%%%%%%%%%%%%%%%%%%%%%%%%%%%%
%%%%%%%%%%%%%%%%%%%%%%%%%%%%%%%%%%%%%%%%%%%%%%%%%%%%%%%%%%%%%%%%%%%%%%%%%%%%%%%%%%%%%%%%%%%%%%%%%%%
\clearpage
\renewcommand{\figurename}{Supplementary Figure}
\renewcommand{\tablename}{Supplementary Table}
\setcounter{section}{0}
\setcounter{figure}{0}
\setcounter{table}{0}

{\LARGE Supplementary Material}

\FloatBarrier

%%%%%%%%%%%%%%%%%%%%%%%%%%%%%%%%%%%%%%%%%%%%%%%%%%%%%%%%%%%%%%%%%%%%%%%%%%%%%%%%%%%%%%%%%%%%%%%%%%%
%%%%%%%%% Human Annotation User Interfaces %%%%%%%%%%%%%%%%%%%%%%%%%%%%%%%%%%%%%%%%%%%%%%%%%%%%%%%%
%%%%%%%%%%%%%%%%%%%%%%%%%%%%%%%%%%%%%%%%%%%%%%%%%%%%%%%%%%%%%%%%%%%%%%%%%%%%%%%%%%%%%%%%%%%%%%%%%%%
\section{Human annotation user interfaces}
\label{sec:human_annotation_user_interfaces}

\begin{figure}[t]
\begin{center}
   \includegraphics[width=0.8\linewidth]{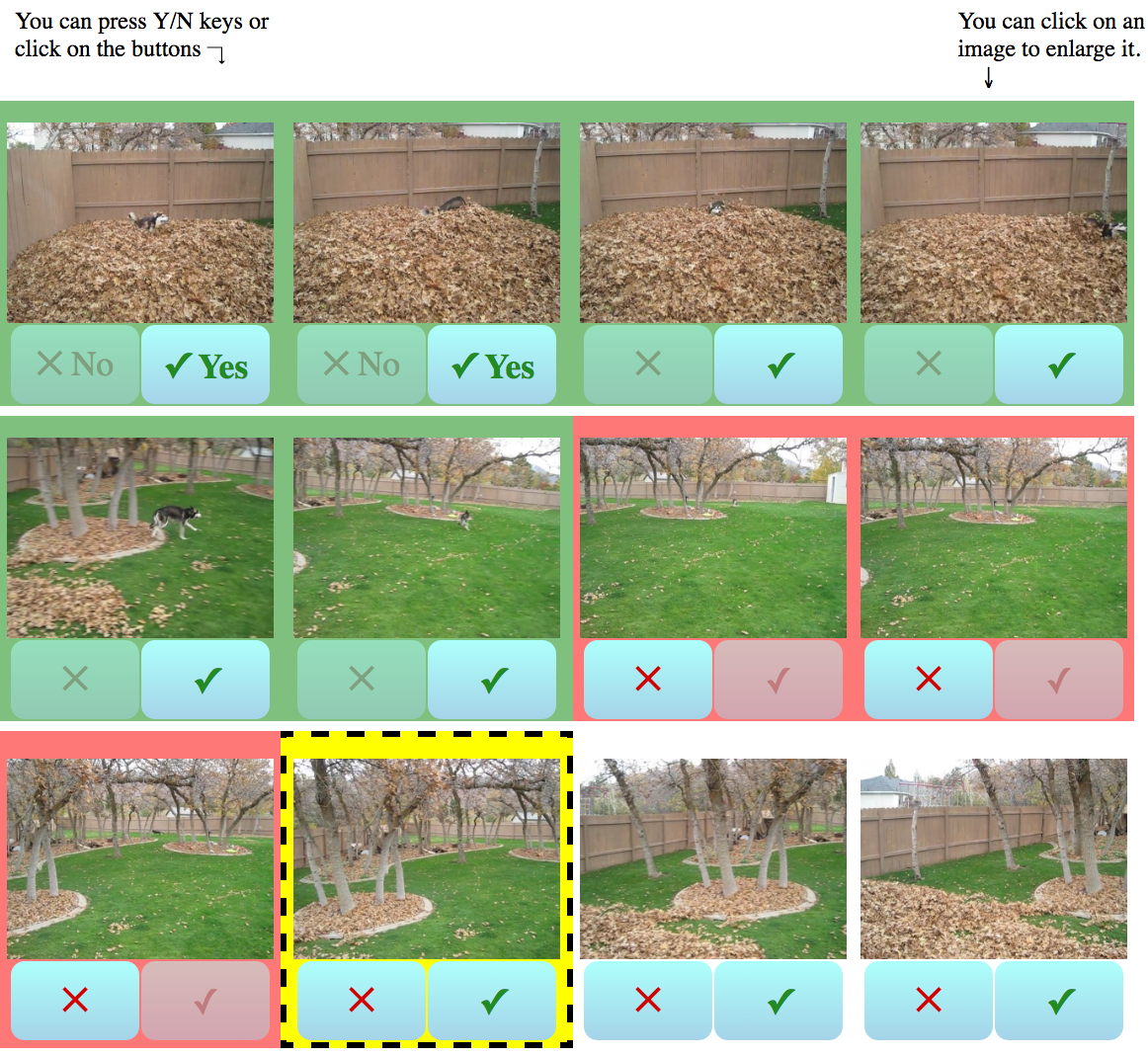}
\end{center}
   \caption{Screenshot of the human annotation tool used to gather frame-level classification labels (stage 2). For each frame, the annotators has to answer whether the class was present or absent. A similar tool was used for segment-level labels (stage 1), displaying fewer frames and allowing only one answer per segment.}
\label{fig:plugin_screenshot_class}
\end{figure}

Supplementary Figure \ref{fig:plugin_screenshot_class} shows a screenshot of the frame-level classification tool. The segment-level tool was very similar. 

\begin{figure}[t]
\begin{center}
   \includegraphics[width=0.8\linewidth]{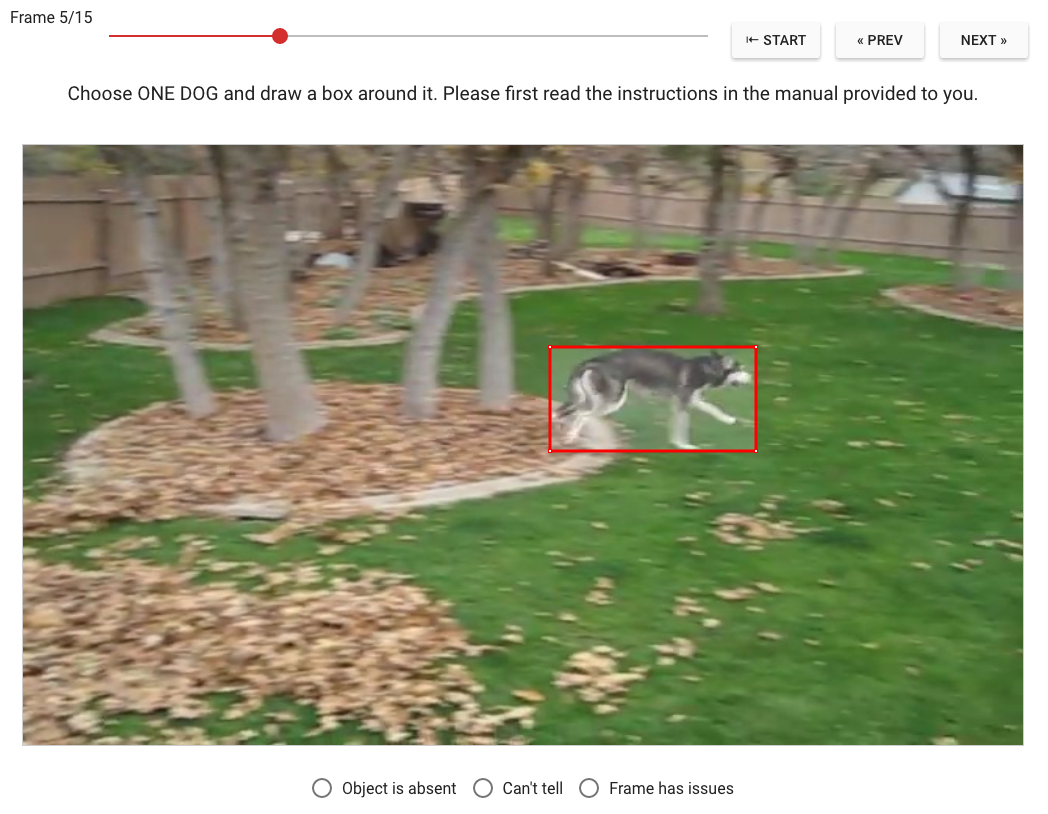}
\end{center}
   \caption{Screenshot of the human annotation tool used to gather bounding boxes (stage 3). At the top, a scroll bar allows navigating through the frames of the segment. The box (red rectagle) can be drawn by clicking and dragging on the image. Three categorical options at the bottom allow the annotator to indicate (i) the absence of the object, (ii) uncertainty, or (iii) problems with the interface. A choice of (i) created an \textit{absent-tag}, which was included in the data set. A choice of (ii) or (iii) resulted in the annotation being discarded. A similar tool was used for the verification stage (stage 4), which instead presented an unchangeable box and an option button to enter the correctness of the box.}
\label{fig:plugin_screenshot_bbox}
\end{figure}

Supplementary Figure \ref{fig:plugin_screenshot_bbox} shows a screenshot of the bounding box drawing tool (stage 3). The bounding box verification tool (stage 4) was similar. In stages 3 and 4, annotators paid careful consideration to object identity: for example, two different dogs in a segment must result in boxes drawn around only one of the dogs. Moreover, all boxes around that one dog must be annotated. For stages 3 and 4, a drawing-verification approach was chosen over a repeated-drawing strategy for two reasons. First, verification is faster this way. Second, having a single drawn box anchors the attention of the verifiers, avoiding problems when multiple instances of the object class are present.

%%%%%%%%%%%%%%%%%%%%%%%%%%%%%%%%%%%%%%%%%%%%%%%%%%%%%%%%%%%%%%%%%%%%%%%%%%%%%%%%%%%%%%%%%%%%%%%%%%%
%%%%%%%%% Attention Span of Human Annotators %%%%%%%%%%%%%%%%%%%%%%%%%%%%%%%%%%%%%%%%%%%%%%%%%%%%%%
%%%%%%%%%%%%%%%%%%%%%%%%%%%%%%%%%%%%%%%%%%%%%%%%%%%%%%%%%%%%%%%%%%%%%%%%%%%%%%%%%%%%%%%%%%%%%%%%%%%
\section{Attention span of human annotators}
\label{sec:attention_span}

In order to gather reliable data, it was necessary to define the classes precisely, so as to avoid too many corner cases. For example, just asking whether an ``airplane'' is present may bring up questions like: ``what if it's a toy airplane?''. Ideally, one would have liked to present the human annotators with the dictionary definition for the class. In practice, however, the attention span of the average untrained, unvetted annotator made this infeasible. In fact, we found that in order to get consistent answers it helped to simplify the questions as much as possible. Some annotators tended to not read the questions completely, even when they consisted of only a handful of lines. This presented a dilemma: on the one hand we needed well-defined classes, on the other hand the questions had to be short. To resolve this dilemma, we opted to split a question into a series of binary choices. Each choice was made by a different rater. Only frames which got a positive result for a given choice made it to the next choice. For example, for the segment-level annotations for the ``airplane'' class, we used the following three choices:
\begin{enumerate}
  \item Can you see the OUTSIDE of a {\bf real airplane} in any frame? Please answer YES even if you cannot see the whole airplane, provided you are confident it is an airplane. Include seaplanes, stealth bombers, \etc.
  \item If the airplane in these frames is:\\
    \textbullet~filmed from the perspective of {\bf someone outside the plane} like a ground observer or someone on another plane
      $\rightarrow$ answer {\bf YES};\\
    \textbullet~filmed from the perspective of {\bf someone inside the plane} like its pilot or a passenger $\rightarrow$ answer {\bf 
      NO}.\\
      If {\bf uncertain}, please answer {\bf NO}.
  \item If the airplane in these frames is:\\
    \textbullet~{\bf REAL} $\rightarrow$ answer {\bf YES};\\
    \textbullet~{\bf NOT REAL} like a {\bf TOY}, cartoon, or {\bf VIDEO GAME} $\rightarrow$ answer {\bf NO}.\\
    If {\bf uncertain or no airplane}, please answer {\bf NO}.
\end{enumerate}

\noindent Notice how some options were structured so that most of the information is at the beginning of the question (``Can you see the outside of a real airplane [...]''). Also, acting on the assumption that annotators read the question only up to the point when they feel they know what it is about, we employed another design principle: structuring the first phrase so that it conveyed zero information until it conveys most of the information. In the example, the phrase ``If the airplane in these frames is filmed from the [...] answer yes'' tells you very little about what the task unless it is read up to the last word. Finally, using caps, bold, and bullets may have helped keep the annotators attention on the text for a bit longer.

%%%%%%%%%%%%%%%%%%%%%%%%%%%%%%%%%%%%%%%%%%%%%%%%%%%%%%%%%%%%%%%%%%%%%%%%%%%%%%%%%%%%%%%%%%%%%%%%%%%
%%%%%%%%% Bounding Box Drawing Guidelines %%%%%%%%%%%%%%%%%%%%%%%%%%%%%%%%%%%%%%%%%%%%%%%%%%%%%%%%%
%%%%%%%%%%%%%%%%%%%%%%%%%%%%%%%%%%%%%%%%%%%%%%%%%%%%%%%%%%%%%%%%%%%%%%%%%%%%%%%%%%%%%%%%%%%%%%%%%%%
\section{Bounding box drawing guidelines}
\label{sec:box_guidelines}

The following rules were observed by annotators during stages 3 and 4.

\begin{itemize}
    \item Objects should be boxed even if only a small part is visible, as long as it is recognizable ({\it airplane example} in figure \ref{fig:main}).
    \item It does not need to be recognizable within the frame in question. The context provided by other frames can be used to deduce the object's identity ({\it train example} in figure \ref{fig:main}).
    \item Only the visible part of the object should be boxed. No inference can take place as to hidden or out-of-frame parts ({\it bear example} in figure \ref{fig:main}).
    \item If an object extends on either side of an occlusion (for example, an elephant behind a narrow tree), one box should be used to include all the visible parts of the object ({\it airplane example} in figure \ref{fig:main}).
    \item The first box is drawn on a random frame within the segment that has a positive classification according to stage 2. (After that, the annotator works forward and backward from that frame.)
\end{itemize}

%%%%%%%%%%%%%%%%%%%%%%%%%%%%%%%%%%%%%%%%%%%%%%%%%%%%%%%%%%%%%%%%%%%%%%%%%%%%%%%%%%%%%%%%%%%%%%%%%%%
%%%%%%%%% Human Annotation Detailed Statistics %%%%%%%%%%%%%%%%%%%%%%%%%%%%%%%%%%%%%%%%%%%%%%%%%%%%
%%%%%%%%%%%%%%%%%%%%%%%%%%%%%%%%%%%%%%%%%%%%%%%%%%%%%%%%%%%%%%%%%%%%%%%%%%%%%%%%%%%%%%%%%%%%%%%%%%%
\section{Human annotation detailed statistics}

Supplementary Tables \ref{tab:classification_counts} and \ref{tab:detection_counts} show the complete counts for all classes for classifications and detections, respectively. Supplementary Table \ref{tab:motion} shows quantitative measures of size and motion for the bounding boxes (next pages).

\begin{table*}
\begin{center}
\begin{tabular}{|l|r|r|r|r|}
\hline
\multicolumn{1}{|c|}{} & \multicolumn{2}{|c|}{Positives} & \multicolumn{2}{|c|}{Negatives} \\
\multicolumn{1}{|c|}{} & \multicolumn{1}{|c|}{Frames} & \multicolumn{1}{|c|}{Videos} & \multicolumn{1}{|c|}{Frames} & \multicolumn{1}{|c|}{Videos} \\
\hline
airplane & 384,448 & 9,314 & 38,847 & 4,446 \\
bear & 354,730 & 7,792 & 57,595 & 5,493 \\
bicycle & 266,317 & 6,352 & 11,121 & 2,348 \\
bird & 476,734 & 11,239 & 49,680 & 5,569 \\
boat & 370,723 & 10,920 & 24,630 & 3,667 \\
bus & 410,742 & 13,800 & 60,724 & 7,473 \\
car & 306,850 & 10,733 & 26,539 & 2,372 \\
cat & 694,265 & 33,019 & 55,281 & 8,582 \\
cow & 465,637 & 17,201 & 69,988 & 9,028 \\
dog & 555,055 & 15,748 & 37,606 & 5,993 \\
elephant & 319,778 & 7,469 & 47,802 & 4,900 \\
giraffe & 49,031 & 1,660 & 8,400 & 1,094 \\
horse & 532,403 & 12,494 & 36,681 & 5,292 \\
knife & 506,180 & 9,563 & 34,256 & 3,518 \\
motorcycle & 338,870 & 12,900 & 24,096 & 4,523 \\
person & 1,810,968 & 79,319 & 132,449 & 21,700 \\
potted plant & 236,509 & 6,940 & 21,326 & 2,766 \\
skateboard & 440,274 & 13,499 & 63,138 & 10,274 \\
toilet & 153,312 & 9,895 & 83,915 & 7,994 \\
train & 339,639 & 11,628 & 76,197 & 5,361 \\
truck & 343,773 & 10,672 & 30,232 & 3,891 \\
umbrella & 189,727 & 7,784 & 25,805 & 4,325 \\
zebra & 26,169 & 1,070 & 7,493 & 823 \\
NONE & 26,457 & 1,589 & \multicolumn{1}{|c|}{--} & \multicolumn{1}{|c|}{--} \\
\hline
ALL & 9,527,784 & 316,235 & 1,021,508 & 128,712 \\
\hline
\end{tabular}
\end{center}
\caption{Human annotation classification counts. We count the number of unique frames and unique videos that have been annotated as having (``positives'') or not having (``negatives'') the class. Due to the fact that we are listing {\it unique} videos, and the fact that occasionally more than one class is annotated per video, the ``ALL'' row is not necessarily the sum of the class rows.}
\label{tab:classification_counts}
\end{table*}

\begin{table*}
\begin{center}
\begin{tabular}{|l|r|r|r|r|}
\hline
\multicolumn{1}{|c|}{} & \multicolumn{2}{|c|}{Bounding Boxes} & \multicolumn{2}{|c|}{Absent Tags} \\
\multicolumn{1}{|c|}{} & \multicolumn{1}{|c|}{Frames} & \multicolumn{1}{|c|}{Videos} & \multicolumn{1}{|c|}{Frames} & \multicolumn{1}{|c|}{Videos} \\
\hline
airplane & 223,712 & 6,932 & 45,319 & 3,621 \\
bear & 231,264 & 6,271 & 31,611 & 3,610 \\
bicycle & 189,955 & 6,122 & 70,911 & 4,168 \\
bird & 228,363 & 8,434 & 42,927 & 4,367 \\
boat & 225,819 & 8,419 & 41,001 & 4,073 \\
bus & 210,565 & 9,132 & 59,121 & 5,670 \\
car & 246,807 & 9,506 & 25,354 & 2,748 \\
cat & 251,472 & 13,828 & 21,867 & 3,882 \\
cow & 197,630 & 10,732 & 73,058 & 7,259 \\
dog & 240,308 & 10,229 & 31,717 & 4,780 \\
elephant & 220,213 & 6,297 & 50,059 & 4,324 \\
giraffe & 42,378 & 1,601 & 10,587 & 1,149 \\
horse & 232,774 & 8,466 & 42,356 & 4,318 \\
knife & 264,296 & 6,837 & 11,785 & 2,127 \\
motorcycle & 223,333 & 10,516 & 48,266 & 4,828 \\
person & 1,285,776 & 68,427 & 283,112 & 36,075 \\
potted plant & 169,260 & 6,036 & 70,349 & 4,889 \\
skateboard & 192,731 & 9,352 & 75,308 & 7,752 \\
toilet & 139,783 & 9,342 & 79,622 & 7,558 \\
train & 239,737 & 8,861 & 45,897 & 3,783 \\
truck & 228,212 & 8,484 & 38,366 & 3,882 \\
umbrella & 114,040 & 5,123 & 90,111 & 6,101 \\
zebra & 20,113 & 1,019 & 7,989 & 782 \\
\hline
ALL & 5,597,399 & 236,102 & 1,291,979 & 129,465 \\
\hline
\end{tabular}
\end{center}
\caption{Human annotation detection counts. We count the number of unique frames and unique videos that have been annotated with bounding boxes (if the object is present) or absent tags. Due to the fact that we are listing {\it unique} videos, and the fact that occasionally more than one object is annotated per video, the ``ALL'' row is not necessarily the sum of the class rows.}
\label{tab:detection_counts}
\end{table*}

\begin{table*}
\begin{center}
\begin{tabular}{|l|c|c|c|c|c|}
\hline
& PF & CF & MA & C-RMS & A-RMS \\
\hline
airplane & 0.86 & 0.80 & 0.43 & 0.094 & 0.103 \\
bear & 0.88 & 0.80 & 0.24 & 0.106 & 0.083 \\
bicycle & 0.72 & 0.65 & 0.24 & 0.138 & 0.092 \\
bird & 0.78 & 0.69 & 0.19 & 0.155 & 0.085 \\
boat & 0.87 & 0.79 & 0.26 & 0.114 & 0.087 \\
bus & 0.80 & 0.73 & 0.42 & 0.086 & 0.123 \\
car & 0.91 & 0.85 & 0.58 & 0.075 & 0.095 \\
cat & 0.92 & 0.84 & 0.44 & 0.115 & 0.121 \\
cow & 0.72 & 0.65 & 0.30 & 0.120 & 0.102 \\
dog & 0.86 & 0.76 & 0.27 & 0.165 & 0.125 \\
elephant & 0.80 & 0.73 & 0.32 & 0.100 & 0.102 \\
giraffe & 0.78 & 0.71 & 0.35 & 0.115 & 0.121 \\
horse & 0.84 & 0.75 & 0.22 & 0.129 & 0.107 \\
knife & 0.96 & 0.89 & 0.33 & 0.122 & 0.126 \\
motorcycle & 0.83 & 0.75 & 0.46 & 0.126 & 0.127 \\
person & 0.80 & 0.70 & 0.25 & 0.122 & 0.096 \\
potted plant & 0.77 & 0.71 & 0.41 & 0.094 & 0.091 \\
skateboard & 0.71 & 0.58 & 0.05 & 0.190 & 0.047 \\
toilet & 0.65 & 0.56 & 0.41 & 0.148 & 0.123 \\
train & 0.87 & 0.81 & 0.50 & 0.072 & 0.111 \\
truck & 0.87 & 0.81 & 0.51 & 0.083 & 0.113 \\
umbrella & 0.78 & 0.70 & 0.37 & 0.122 & 0.123 \\
zebra & 0.67 & 0.60 & 0.33 & 0.119 & 0.122 \\
\hline
\end{tabular}
\end{center}
\caption{Measures of object motion. Each value is an average over all the segments for the corresponding class. {\it Present Fraction (PF)}: fraction of the segment frames in which the object is present. {\it Continuous Fraction (CF)}: fraction of the frames in the longest sequence in which the object was continuously present. This is an indication of how often the object enters and leaves the field of view. {\it Mean Area (MA)}: mean area of the box. {\it Center RMS (C-RMS)}: root-mean-square of the distances the center of the box travels from each frame to the next. This is a measure of sideways object/camera motion. {\it Area RMS (A-RMS)}: root-mean-square of the change in area from each frame to the next. This is an indication of the amount of depth-wise object/camera motion. {\it For all}: areas and distances are measured in the relative coordinate system in which both axes run from 0 to 1, regardless of the aspect ratio of the video. Distances and area changes were only measured over contiguous frames. Everything is based on data at 1 frame per second. Note that there may be significant motion not captured by these quantities, such as: (i) relative movement ``in place'' like in the case of a spinning wheel, (ii) movement of the background, as would be seen when a racecar is kept well centered in the field of view but the background ``passes by'', or (iii) movement of an object that spans the field of view such as a train passing by while a steady camera only captures one wagon at a time.}
\label{tab:motion}
\end{table*}

%%%%%%%%%%%%%%%%%%%%%%%%%%%%%%%%%%%%%%%%%%%%%%%%%%%%%%%%%%%%%%%%%%%%%%%%%%%%%%%%%%%%%%%%%%%%%%%%%%%
\section{Relevant GitHub locations}
%%%%%%%%%%%%%%%%%%%%%%%%%%%%%%%%%%%%%%%%%%%%%%%%%%%%%%%%%%%%%%%%%%%%%%%%%%%%%%%%%%%%%%%%%%%%%%%%%%%
\label{github_section.}
The following are locations for related GitHub models: \\
Inception-v3: \\
https://github.com/tensorflow/models/tree/master/slim \\
Inception-ResNet-v2: \\
https://github.com/tensorflow/models/tree/master/slim \\
Faster-RCNN: \\
https://github.com/rbgirshick/py-faster-rcnn \\

%%%%%%%%%%%%%%%%%%%%%%%%%%%%%%%%%%%%%%%%%%%%%%%%%%%%%%%%%%%%%%%%%%%%%%%%%%%%%%%%%%%%%%%%%%%%%%%%%%%
%%%%%%%%% Per-class statistics of object detection baseline %%%%%%%%%%%%%%%%%%%%%%%%%%%%%%%%%%%%%%%
%%%%%%%%%%%%%%%%%%%%%%%%%%%%%%%%%%%%%%%%%%%%%%%%%%%%%%%%%%%%%%%%%%%%%%%%%%%%%%%%%%%%%%%%%%%%%%%%%%%

\section{Per-class object detection baseline}

Supplementary Table \ref{tab:perclass} shows the difficulty of object detection for each class (next pages).

\begin{table*}
\begin{center}
\begin{tabular}{ |c|c|c|c|c| }
\hline
& \multicolumn{2}{c|}{COCO model} & \multicolumn{2}{|c|}{YT-BB model} \\
\hline
\textit{eval on:} & COCO & YT-BB & COCO & YT-BB \\
\hline
airplane& 0.56 & 0.65 & 0.41 & 0.72 \\
bear& 0.80 & 0.45  & 0.66 & 0.68 \\
bicycle & 0.27 & 0.15 & 0.14 & 0.40 \\
bird & 0.17 & 0.29 & 0.15 & 0.45 \\
boat & 0.17 & 0.23 & 0.09 & 0.47 \\
bus & 0.56 & 0.61 & 0.47 & 0.77 \\
car & 0.29 & 0.43 & 0.06 & 0.81 \\
cat & 0.72 & 0.49 & 0.61 & 0.62 \\
cow & 0.40 & 0.38 & 0.25 & 0.59 \\
dog & 0.58 & 0.29 & 0.48 & 0.52 \\
elephant & 0.58 & 0.55 & 0.50 & 0.67 \\
giraffe & 0.80 & 0.77 & 0.54 & 0.67 \\
horse & 0.63 & 0.40 & 0.41 & 0.56 \\
knife & 0.10 & 0.16 & 0.02 & 0.60 \\
motorcycle & 0.35 & 0.48 & 0.26 & 0.59 \\
person & 0.41 & 0.12 & 0.23 & 0.41 \\
potted plant & 0.21 & 0.15 & 0.08 & 0.39 \\
skateboard & 0.42 & 0.41 & 0.22 & 0.43 \\
toilet & 0.68 & 0.60 & 0.29 & 0.71 \\
train & 0.49 & 0.58 & 0.55 & 0.73 \\
truck & 0.29 & 0.38 & 0.14 & 0.71 \\
umbrella & 0.26 & 0.29 & 0.11 & 0.55 \\
zebra & 0.56 & 0.87 & 0.49 & 0.59 \\
\hline
\end{tabular}
\end{center}
\caption{Measured difficulty of object detection for each class trained on the COCO and YT-BB data sets. All values are calculations of the mean average precision (mAP) across precision-recall curves. Each column indicates evaluation on COCO and YT-BB, respectively.}
%\caption{Measured difficulty of object detection for each class trained on the COCO (left table) and YT-BB (right table) data sets. All values are mean average precision (mAP) across precision recall curves. Left and right columns indicate evaluation on COCO and YT-BB, respectively.}
\label{tab:perclass}
\end{table*}

\end{document}